\documentclass[11pt]{article}

\usepackage[final]{acl}

\usepackage{algorithm}

\usepackage{times}
\usepackage{latexsym}
\usepackage[noend]{algpseudocode}
\usepackage{pifont}
\usepackage{amsmath}
\usepackage{bbding}
\usepackage{xcolor}
\usepackage{booktabs}

\usepackage{listings}
\usepackage[T1]{fontenc}
\usepackage[utf8]{inputenc}

\usepackage{microtype}

\usepackage{inconsolata}

\usepackage{graphicx}

\title{VLN-NF: Feasibility-Aware Vision-and-Language Navigation with False-Premise Instructions}
\author{
Hung-Ting Su\textsuperscript{1}\thanks{This work was conducted at National Taiwan University. The author is currently affiliated with DRIC.}\thanks{Project lead.}\thanks{Equal technical contribution.}
\quad
Ting-Jun Wang\textsuperscript{1}\footnotemark[3]
\quad
Jia-Fong Yeh\textsuperscript{1}
\quad
Min Sun\textsuperscript{2}
\quad
Winston H. Hsu\textsuperscript{1} \\ \\
\textsuperscript{1}National Taiwan University
\quad
\textsuperscript{2}National Tsing Hua University
\\
 \url{https://vln-nf.github.io/}
}

\begin{document}
\maketitle

\newcommand{\OurDataset}{\textsc{VLN-NF}}
\newcommand{\OurDatasetFull}{VLN-Not Found}
\newcommand{\OurModel}{ROAM}
\newcommand{\OurModelFull}{Room-Object Aware Movement}
\newcommand{\OurMetric}{REV-SPL}
\newcommand{\OurMetricFull}{Reach--Explore--Verify SPL}
\begin{abstract}
Conventional Vision-and-Language Navigation (VLN) benchmarks assume instructions are feasible and the referenced target exists, leaving agents ill-equipped to handle false-premise goals. We introduce \textbf{\OurDataset{}}, a benchmark with false-premise instructions where the target is absent from the specified room and agents must navigate, gather evidence through in-room exploration, and explicitly output \texttt{NOT-FOUND}. \OurDataset{} is constructed via a scalable pipeline that rewrites VLN instructions using an LLM and verifies target absence with a VLM, producing plausible yet factually incorrect goals. We further propose \textbf{\OurMetric{}} to jointly evaluate room reaching, exploration coverage, and decision correctness. To address this challenge, we present \textbf{ROAM}, a two-stage hybrid that combines supervised room-level navigation with LLM/VLM-driven in-room exploration guided by a free-space clearance prior. ROAM achieves the best \OurMetric{} among compared methods, while baselines often under-explore and terminate prematurely under unreliable instructions. %\OurDataset{} project page can be found at \url{https://vln-nf.github.io/}.

%Conventional VLN benchmarks largely assume that instructions are feasible and that the referenced target exists, which leaves agents ill-equipped to handle false-premise goals. We introduce VLN-NF, a VLN benchmark where the instruction may refer to a plausible target that is absent from the specified room, and agents must (i) navigate to the relevant room, (ii) explore to gather evidence, and (iii) explicitly output NOT‑FOUND when the target cannot be located. 
%VLN‑NF is constructed with a scalable pipeline that pairs each feasible REVERIE episode with an infeasible counterpart: an LLM rewrites the target entity while preserving surrounding context, and a VLM checks that the new target is not present in the target-room panoramas. %This process yields linguistically natural instructions with factually incorrect goal entities; a human audit on a subset indicates <2\% generation errors. 
%Because exploration is underdetermined under partial observability, standard VLN metrics can reward premature termination. 
%Also, we propose REV‑SPL, which jointly evaluates room reaching, in-room coverage, and the final FOUND/NOT‑FOUND decision, while penalizing both inefficient wandering and unsupported early abstention. Finally, we present ROAM, a two-stage approach that combines weakly supervised room localization with LLM/VLM-driven in-room exploration guided by a free-space clearance prior. Across supervised and LLM-based baselines, ROAM achieves the best overall performance on VLN‑NF, highlighting the need for evidence-grounded failure reporting in VLN.
\end{abstract}

\section{Introduction}

Language provides a flexible interface for specifying goals to situated agents in the physical world. Vision-and-Language Navigation (VLN) is a crucial robotic task studying language grounding and path planning under partial observability. Existing VLN datasets \cite{anderson2018vision,ku2020room,qi2020reverie,thomason2020vision} and supervised agents \cite{pashevich2021episodic,chen2021history,chen2022think} or agents based on LLM built on them \cite{zhou2024navgpt,zhou2024navgpt2,chen2024mapgpt,lin2025navcot} typically assume that \emph{every instruction is feasible}. In real deployments, however, humans often make mistakes when instructing robots. For example, a user may say ``\emph{pick up the plate on the table in the kitchen}" when the plate is actually elsewhere (e.g., in the lounge or even in the car). A cognitive science study \cite{wang2002study} found that humans mislocate an item roughly once in every seven object-location recall trials. Such instructions are semantically well-formed but factually incorrect, and they can cause agents to hallucinate similar objects or search indefinitely. Therefore, an agent must be able to explore and explicitly report when the target cannot be found, rather than simply assuming success and exploiting task priors.

\begin{figure*}[t]
  \includegraphics[width=\textwidth]{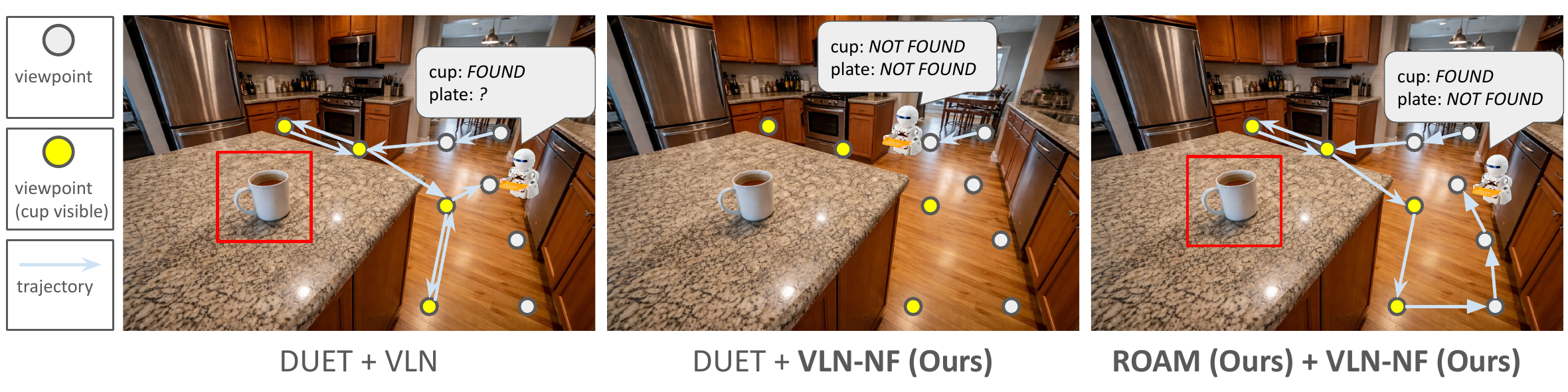}
 \caption{
\textbf{A toy illustration of failure modes under unreliable instructions.}
For compactness, we visualize two \emph{separate single-target} queries in the same scene (cup: feasible; plate: infeasible/absent in the kitchen).
\textbf{Left: Standard VLN (DUET+VLN)} lacks an explicit \texttt{NOT-FOUND} output (hence ``?'' for plate). 
\textbf{Middle: Adding \texttt{NOT-FOUND} to action space (DUET+VLN-NF)} can lead to premature abstention. 
\textbf{Right: Our proposed ROAM+VLN-NF} performs evidence-gathering exploration and outputs the correct decisions. 
}

  \label{fig:fig1}
\end{figure*}

Related efforts study instruction unreliability across embodied settings, including feasibility in 2D interfaces~\cite{burns2022motif}, stepwise VLN instruction errors~\cite{taioli2024mind}, obstruction-induced instruction--environment mismatch~\cite{hong2024r2runo}, and false-premise/abstention/hallucination in embodied QA or VLA~\cite{wu2024noisyeqa,wu2025abstaineqa,hsieh2025dowhat,chakraborty2025heal}. 
Yet \emph{evidence-grounded} \texttt{NOT-FOUND} for \emph{3D partially observable} VLN---where absence must be justified through autonomous exploration---is still largely unaddressed. 
To this end, we introduce \textbf{\OurDatasetFull{} (\OurDataset{})}, a new VLN task that evaluates agents' ability to handle false-premise instructions referring to non-existent targets. 
We focus on this setting as a controlled but practical form of goal-level false premise, where absence must be established through exploration under partial observability rather than inferred from a single observation. 
In \OurDataset{}, the agent must predict \texttt{NOT-FOUND} when the target object is absent from the region specified by the instruction, indicating that the instruction is infeasible. 
To enable low-cost construction, we develop a scalable pipeline that converts feasible VLN instances into infeasible ones via instruction rewriting and automatic verification. An LLM replaces the original target with a plausible but absent alternative to simulate false-premise instructions, and a VLM verifies the target's absence. This yields linguistically natural instructions with factually incorrect target descriptions. Human evaluation on a subset confirms high quality, with \textless{}2\% errors.

Unlike conventional VLN, where success is defined by reaching a single goal and efficiency is measured against a shortest path, \OurDataset{} requires \emph{evidence-gathering exploration} under partial observability: the absence of a target cannot be confirmed from a single viewpoint. Naively reusing standard VLN metrics encourages degenerate behaviors (e.g., stopping early or predicting \texttt{NOT-FOUND} without sufficient search). To enable principled comparison, we refine SPL and propose \textbf{\OurMetricFull{} (\OurMetric{})}. \OurMetric{} (1) defines a reference exploration protocol that specifies \emph{where} an agent should reasonably search given the instruction (e.g., local exploration when landmark cues exist and a broader room sweep otherwise), and uses the resulting reference length to normalize exploration efficiency; (2) evaluates decision quality on mixed feasible/infeasible episodes, capturing the trade-off between abstaining when appropriate and avoiding false alarms; and (3) discourages premature termination by rewarding coverage of the target room and penalizing unsupported \texttt{NOT-FOUND}. Together, this design extends SPL from shortest-path goal reaching to evidence-grounded verification under partial observability.

Standard VLN benchmarks and agents typically assume feasibility and do not expose an explicit \texttt{NOT-FOUND} decision (Fig.~\ref{fig:fig1}, left). 
Naively adding a \texttt{NOT-FOUND} action and training via imitation learning on reference exploration trajectories can in turn encourage premature abstention (Fig.~\ref{fig:fig1}, middle). 
This arises because exploration is underdetermined—many trajectories can be reasonable and there is no explicit ``search finished'' signal—so imitation-learned VLN agents can suffer covariate shift \cite{Kamath_2023_CVPR,seo2024drildice} and compounding errors \cite{zhang2024diffusionmeetsdagger}, which often manifest as unreliable stopping behavior \cite{xiang-etal-2020-learning}. 
In \OurDataset{}, this can surface as premature \texttt{NOT-FOUND} predictions that are not supported by sufficient evidence \cite{chakraborty2025heal,wu2025abstaineqa}. 
Moreover, purely supervised approaches typically require costly trajectory-level annotations (especially for exploration), while LLM-based agents often struggle with room-to-room navigation under partial observability without stepwise guidance. 
To tackle this challenge, we propose a two-stage hybrid framework, \textbf{R}oom-\textbf{O}bject \textbf{A}ware \textbf{M}ovement (\textbf{ROAM}; Fig.~\ref{fig:fig1}, right), which combines supervised room localization with LLM/VLM-driven in-room exploration and verification. 
ROAM first localizes the target room using a weakly supervised model, then evidence-gathering explores the room and decides whether the target object is present or should be reported as \texttt{NOT-FOUND}.

Experimental results show that existing VLN agents struggle on \OurDataset{}: a supervised baseline \cite{chen2022think} achieves 4.2 REV-SPL, while LLM-based agents \cite{zhou2024navgpt,georgiev2024gemini} reach only 1.0 and 1.5 REV-SPL. In contrast, \textbf{ROAM} attains 6.1 REV-SPL, improving over the supervised baseline by 45\% while requiring only room-level annotations.

\paragraph{Contributions.} \textbf{1. Task, dataset, and construction.} We introduce \textbf{\OurDatasetFull{} (\OurDataset{})}, a VLN benchmark of false-premise instructions requiring evidence-gathering exploration and \texttt{NOT-FOUND}, built via an LLM rewrite + VLM absence verification pipeline with \textless{}2\% human-judged errors.
\textbf{2. Evaluation.} We propose \textbf{REV-SPL}, a variant of SPL that uses a reference exploration protocol and decision-quality metrics to discourage premature stopping and unsupported \texttt{NOT-FOUND}.
\textbf{3. Method.} We present \textbf{ROAM}, a two-stage hybrid that localizes the target room with supervision and then performs VLM/LLM-driven in-room exploration and verification.
\textbf{4. Results.} ROAM achieves \textbf{6.1} REV-SPL, outperforming a supervised baseline by \textbf{45\%} while requiring only room-level annotations.

\section{Related Work}

\textbf{Vision-and-Language Navigation.}
VLN studies how an embodied agent follows natural-language instructions to navigate in partially observed 3D environments.
Benchmarks such as R2R~\cite{anderson2018vision}, RxR~\cite{ku2020room}, REVERIE~\cite{qi2020reverie}, and CVDN~\cite{thomason2020vision} (built on Matterport3D~\cite{chang2017matterport3d}) have driven rapid progress, but they typically assume that the instruction is \emph{feasible} and do not require agents to explicitly report failure.
A related challenge is reliable exploration and termination under partial observability: exploration is often underdetermined and stopping decisions can be brittle, especially under imitation learning due to covariate shift and compounding errors~\cite{Kamath_2023_CVPR,seo2024drildice,zhang2024diffusionmeetsdagger,xiang-etal-2020-learning}.
\OurDataset{} targets this reliability axis by requiring evidence-grounded \texttt{NOT-FOUND} in addition to navigation.

\textbf{VLN agents and instruction augmentation.}
VLN approaches span memory-centric models~\cite{pashevich2021episodic,chen2021history}, map-based/topological methods~\cite{chen2022think,wang2023gridmm,georgakis2022cross}, and LLM-based planners or copilots~\cite{zhou2024navgpt,zhou2024navgpt2,pan2023langnav,qiao2024llm,long2024discuss,chen2024mapgpt,lin2025navcot}.
Several works further improve navigation by rewriting or generating auxiliary guidance, e.g., disambiguating landmarks~\cite{zhang2024navhint,zhang2023vln}, jointly learning instruction following and generation~\cite{wang2023lana}, or enabling controllable instruction generation~\cite{kong2024controllable}.
These methods largely assume that a valid goal exists and use rewriting to make instructions \emph{easier to follow}.
In contrast, our pipeline rewrites feasible instances into \emph{false-premise} (infeasible) ones and automatically verifies target absence, enabling systematic evaluation of \texttt{NOT-FOUND} decision-making.

\textbf{Instruction unreliability in VLN.}
Prior work has examined how corrupted or imperfect instructions affect VLN agents.
\citet{hahn2023way} perturb instruction tokens (e.g., directions, nouns, numbers) to analyze sensitivity, highlighting the importance of object nouns.
Mind the Error~\cite{taioli2024mind} modifies VLN-CE instructions~\citep{krantzvlnce2020} and studies detection/correction of \emph{stepwise} instruction errors, where the high-level goal remains valid but intermediate guidance is flawed.
Our setting differs in that the instruction can be semantically well-formed yet factually incorrect at the \emph{goal level} (false premise), so the correct behavior is to gather evidence and output \texttt{NOT-FOUND} when appropriate.

\textbf{Infeasibility, mismatch, and abstention across embodied tasks.}
MoTIF~\cite{burns2022motif} introduces unknown command feasibility with follow-up questions, but operates in a mobile-app setting where the agent has a fully observable 2D screen and access to a complete human exploration trace.
R2R-UNO~\cite{hong2024r2runo} creates instruction--environment mismatch via unexpected obstructions, focusing on navigability changes rather than semantic goal absence.
CARe~\cite{su2024contextaware} revises object-navigation decisions when pre-explored semantic maps are inaccurate using uncertainty estimation and multi-view consistency.
Concurrent with our work, ADAPT~\cite{chen2026adaptbenchmarkingcommonsenseplanning} also studies infeasibility during navigation, but focuses more on resource-level affordance failures (e.g., unavailable ovens) than on evidence-grounded absence verification.
Beyond navigation, recent work studies false-premise handling in VLA manipulation~\cite{hsieh2025dowhat} and abstention or hallucinations under instruction--visual inconsistencies in embodied QA and embodied agents~\cite{wu2024noisyeqa,wu2025abstaineqa,chakraborty2025heal}.
\textbf{\OurDataset{} complements these lines by instantiating \emph{goal-level false premises} in 3D partially observable VLN, where absence must be justified through autonomous exploration and explicit \texttt{NOT-FOUND}.}

\section{The \OurDataset{} Dataset}\label{sec:dataset}

\OurDataset{} builds on the Matterport3D environment~\cite{chang2017matterport3d} and extends REVERIE~\cite{qi2020reverie} with infeasible goal descriptions. 
We adopt REVERIE as our base benchmark because it emphasizes goal-driven navigation and remote object grounding: instructions typically specify the target (and its surrounding context) rather than prescribing a step-by-step route as in R2R~\cite{anderson2018vision} and related datasets. 
Importantly, our construction pipeline is modular and can be transferred to other VLN datasets in future work. 

%\OurDataset{} contains 4,724 training instructions (retaining $\sim$45\% of REVERIE after filtering those incompatible with \OurDataset{}) across 55 scenes, 468 validation instructions in seen environments (38 scenes), and 1,436 validation instructions in unseen environments (10 scenes).
%On average, each instruction has 19.1 words, and the dataset spans 1,255 distinct target object types, enabling evaluation across diverse categories and spatial references under instruction uncertainty.

\begin{table}[t]
\centering
\small
\setlength{\tabcolsep}{5pt}
\begin{tabular}{lrrrrr}
\hline
Split & \#Scans & \#Pairs & \#Instr. & \#\texttt{FOUND} & \#NF \\
\hline
Train     & 55 & 2,362 & 4,724 & 2,362 & 2,362 \\
Val-seen  & 38 &   234 &   468 &   234 &   234 \\
Val-unseen& 10 &   718 & 1,436 &   718 &   718 \\
\hline
\end{tabular}
\caption{\textbf{\OurDataset{} statistics.} We use \emph{scan} to denote a Matterport3D building (scan ID).
Each \emph{pair} consists of one original REVERIE instruction (\texttt{FOUND}, feasible) and one generated \OurDataset{} instruction (NF, \texttt{NOT-FOUND}/infeasible), sharing the same scan, start viewpoint, and target room.
Val-seen scans are a subset of the training scans.}
\label{tab:dataset_stats}
\end{table}

\paragraph{Dataset statistics and pairing.}
Table~\ref{tab:dataset_stats} summarizes \OurDataset{}.
We inherit REVERIE's environment, viewpoint graph, and official train/val\_seen/val\_unseen splits~\cite{qi2020reverie} (Matterport3D scans~\cite{chang2017matterport3d}); we use \emph{scan} to denote a building and \emph{room} to denote a segmented region within a scan. 
Starting from REVERIE, we filter out instruction instances that are incompatible with \OurDataset{}, including:
(i) instructions without a well-defined target room/object for reliable navigation and verification,
(ii) cases where our rewriting+verification pipeline cannot confidently produce an absent-but-plausible replacement target after resampling (e.g., the verifier repeatedly detects the substituted object), and
(iii) episodes where the target-room connectivity/visibility is insufficient for our reference exploration protocol to achieve the minimum object-coverage requirement (used for filtering and evaluation).
For each remaining REVERIE instruction instance, we construct a one-to-one \OurDataset{} counterpart by rewriting the target object to a plausible but absent object and verifying its absence, yielding paired \texttt{FOUND} (feasible)/NF (\texttt{NOT-FOUND}, infeasible) instructions with identical navigation context (same scan, start viewpoint, and target room). The detailed statistics are shown in Table~\ref{tab:dataset_stats}.
%As shown in Table~\ref{tab:dataset_stats}, \OurDataset{} contains 4,724 training instruction instances across 55 training scans, and a balanced feasible/infeasible mixture in every split.
%Val\_seen contains episodes from 38 scans that overlap with training (hence fewer scans than train), while val\_unseen uses 10 disjoint scans.
%Overall, instructions contain 19.1 words on average and cover 1,255 distinct target-object surface forms, enabling evaluation under realistic instruction uncertainty.

\begin{figure*}[t]
    \centering
    \includegraphics[width=\textwidth]{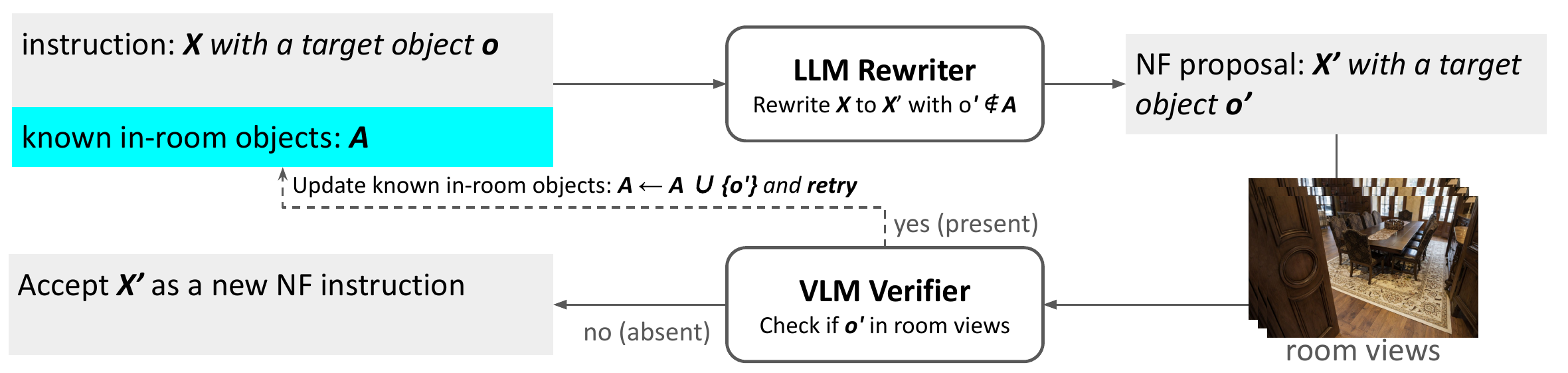}
    %{latex/images/figure_2.png}
    \caption{
We design a scalable pipeline to rewrite VLN instructions and generate NF (\texttt{NOT-FOUND}, infeasible) ones. 
An LLM-based rewriter uses commonsense priors to generate physically plausible instructions (e.g., \textit{water the plant underneath the window}→\textit{clean the couch underneath the window}), while a VLM-based verifier checks that the referenced target entity is absent from the scene, ensuring NF correctness (Sec.~\ref{sec:dataset}).
}
    %Our instruction adjustment pipeline leverages an LLM and a Multimodal LLM—referred to as the Rewriter and the Verifier, respectively—to generate high-quality unreliable instructions that simulate real-world human errors.
    
    \label{fig:figure2}
\end{figure*}

\subsection{Task Definition}\label{subsec:task_definition}
\OurDataset{} extends REVERIE-style VLN~\cite{qi2020reverie} with a \texttt{NOT-FOUND} option.
An agent follows an instruction $X=\langle w_1,\dots,w_L\rangle$ to navigate a Matterport3D scan represented as a graph $\mathcal{G}=(\mathcal{V},\mathcal{E})$, where each node $v\in\mathcal{V}$ is a navigable viewpoint with a panoramic RGB observation (discretized into $n$ view images).
At step $t$, the agent selects (i) a neighboring viewpoint to move to, (ii) \textsc{STOP} and output a grounded target object as in REVERIE, or (iii) \texttt{NOT-FOUND} to abstain when the instruction is infeasible (i.e., the referenced target does not exist in the specified room).
Thus, each episode requires both navigation to the target room and a final decision of \texttt{FOUND} vs.\ \texttt{NOT-FOUND}.

\subsection {Dataset Construction} \label{subsec:dataset_construction}

%To mirror real-world scenarios where instructions specify only the target rather than providing step-by-step guidance as in R2R \cite{anderson2018vision}. We build \OurDataset{} on the REVERIE dataset \cite{qi2020reverie}, which requires agents to navigate and identify remote objects in real indoor environments. We extend the dataset by generating infeasible instructions, where the described object does not exist in the specified location. As a result, half of the instructions come from the original REVERIE dataset, while the other half are generated by our pipeline. This is done by rewriting the original instruction and verifying the infeasibility of the new reference, as illustrated in Figure~\ref{fig:figure2}. We show an example from our dataset generated by this data generation pipeline in the appendix, listing 2.
Annotating exploration behavior is inherently challenging: exploration trajectories are typically long (making dense annotation costly) and underdetermined—many distinct paths can be valid—unlike traditional VLN exploitation, where shortest paths provide a clear reference. To address this, we propose a scalable pipeline to augment existing VLN datasets by (1) rewriting each instruction to create a linguistically natural but infeasible counterpart, (2) constructing an exploration protocol that can supervise imitation learning and serve as reference labels for evaluation. %In addition, we introduce metrics that assess exploration quality from complementary perspectives (e.g., coverage, decision correctness, and efficiency).

As shown in Fig.~\ref{fig:figure2}, we generate an infeasible counterpart for each REVERIE instance via a \textbf{rewrite-and-verify} loop\footnote{An example is provided in the Appendix.}. 
Given instruction $\mathcal{X}$ with target object $o$, the Rewriter \footnote{Prompts are provided in the Appendix.} proposes a plausible replacement $o'$ that is not in the target-room object list 
%\footnote{For datasets without room-level object lists, $\mathcal{A}$ can be initialized as empty and iteratively expanded by the verifier feedback described below.} 
$\mathcal{A}$, and rewrites $\mathcal{X}$ into $\mathcal{X}'$ by substituting $o\!\rightarrow\!o'$ while enforcing commonsense plausibility. 
The Verifier then runs open-vocabulary detection over all panoramas in the target room; if any view detects $o'$, we blacklist it and resample, otherwise we accept $\mathcal{X}'$ as an NF instruction. 
A manual audit on 5\% of pairs yields $<2\%$ errors\footnote{Details are provided in the Appendix.}, indicating that our rewriter-verifier pipeline reliably produces truly infeasible (NF) targets in the specified room.

\begin{algorithm}[t]
\caption{Extract Reach Path 
%(Prefix Until Entering Target Room)
}
\label{alg:reach_path}
\small
\begin{algorithmic}[1]
\State $\text{original\_final\_vp} \gets \text{gt\_path}[-1]$
\State $\text{room\_vps} \gets \textsc{VpsInSameRoom}(\text{original\_final\_vp})$
\State $\text{reach\_path} \gets [\ ]$
\For{each $\text{vp}$ in $\text{gt\_path}$}
    \State $\text{reach\_path.append}(\text{vp})$
    \If{$\text{vp} \in \text{room\_vps}$}
        \State \textbf{break}
    \EndIf
\EndFor
\State \Return $\text{reach\_path}$
\end{algorithmic}
\end{algorithm}

\begin{algorithm}[t]
\caption{Reference Exploration Path (Localized / Landmark-Cued)}
\label{algo:haslm}
\small
\begin{algorithmic}[1]
\State $o$: original target object (from REVERIE annotations)
\State $\mathcal{V}_{\text{vis}}(o)$: viewpoints in the target room that can see $o$ 
%(from BBox visibility)
\State $v_{\text{start}}$: endpoint of the reach path (room entry viewpoint)
\Statex
\State $\text{cand\_vps} \gets \mathcal{V}_{\text{vis}}(o)$
\State $\text{explore\_path} \gets \textsc{BruteForceTSP}(v_{\text{start}}, \text{cand\_vps})$
\State \Return $\text{explore\_path}$
\end{algorithmic}
\end{algorithm}

\begin{algorithm}[t]
\caption{Reference Exploration Path (Coverage / No-Landmark)}
\label{algo:nolm}
\small
\begin{algorithmic}[1]
\State $v_{\text{start}}$ = endpoint of the reach path (room entry viewpoint)
\State $\mathcal{V}_{\text{room}}$ = all viewpoints in the target room
\State $\mathcal{O}$ = set of all objects in the target room
\State $\mathcal{O}_{\text{seen}} = \emptyset$
\State $\text{explore\_path} = [v_{\text{start}}]$
\State $\text{visited} = \{v_{\text{start}}\}$
\While{True}
    \State $v_{\text{curr}} = \text{explore\_path}[-1]$
    \State $\mathcal{N} = \text{unvisited neighbors of } v_{\text{curr}}$
    \If{$\mathcal{N} = \emptyset$}
        \State \textbf{break}
    \EndIf
    \State $v_{\text{next}} = \arg\max\limits_{v \in \mathcal{N}} |\textsc{ObjectsVisible}(v)\setminus \mathcal{O}_{\text{seen}}|$
    \State $\text{explore\_path.append}(v_{\text{next}})$
    \State $\text{visited.add}(v_{\text{next}})$
    \State $\mathcal{O}_{\text{seen}} \gets \mathcal{O}_{\text{seen}} \cup \textsc{ObjectsVisible}(v_{\text{next}})$
    \If{$|\mathcal{O}_{\text{seen}}| / |\mathcal{O}| \geq 1.0$}
        \State \textbf{break}
    \EndIf
\EndWhile
\If{$|\mathcal{O}_{\text{seen}}| / |\mathcal{O}| < 0.85$}
    \State \textbf{discard} the instruction-path pair
\Else
    \State \Return $\text{explore\_path}$
\EndIf
\end{algorithmic}
\end{algorithm}

\paragraph{Generating Reference Exploration Paths.}
To decide \texttt{FOUND} vs.\ \texttt{NOT-FOUND}, an agent must explore the target room; however, exploration trajectories are long and underdetermined, making dense human annotation costly.
We therefore define a \emph{reference exploration protocol} (used for training/evaluation only; not exposed to the agent) with two cases.
(1) \textbf{Localized / landmark-cued search.} When the instruction provides strong spatial cues (e.g., ``beside the chair''), the plausible search region is localized.
Since our NF instruction is created by replacing the target object while keeping the surrounding description unchanged, we approximate this region by the viewpoints in the target room that can see the original target object $o$. 
The candidate set is small (typically $N\!\le\!15$), so we compute a shortest visit order starting from the room entry viewpoint using an exhaustive TSP solver (Alg.~\ref{algo:haslm}).
(2) \textbf{Coverage / no-landmark search.} Without such cues, we use a greedy explorer that repeatedly moves to the neighboring viewpoint that reveals the most \emph{new} room objects , stopping when all room objects are observed or no unvisited neighbors remain (Alg.~\ref{algo:nolm}).
If the final object coverage is below 85\%, we discard the instance. Across retained subsets induced by stricter minimum reference-coverage thresholds (0.85/0.90/0.95), method ranking remains unchanged; we report the full sensitivity analysis in the appendix.
% -------------------- End reference paths --------------------

%\subsection {Human evaluation}

\subsection{Evaluation Metrics}\label{subsec:evaluation_metrics}
Exploration under partial observability is underdetermined and long-horizon, so metrics must avoid rewarding premature ``I'm done'' decisions.
In \OurDataset{}, an agent must (i) \emph{reach} the target room and (ii) make a final \emph{decision} of \texttt{FOUND} vs.\ \texttt{NOT-FOUND}.
We report a primary composite metric (REV-SPL) together with a small set of decomposed diagnostics for interpretability.

\paragraph{Success Rates.}
\textbf{Reach SR} is the fraction of episodes where the agent visits at least one viewpoint inside the target room. 
%\textbf{Decision F1} measures decision quality on the mixed feasible/infeasible set: predicting \texttt{NOT-FOUND} on infeasible episodes and \textsc{FOUND} on feasible episodes (with the same grounding criterion as REVERIE).
\textbf{Reach\&Decision SR} counts episodes where the agent both reaches the correct room and makes the correct final decision. 

\paragraph{Path-length Efficiency.}
Following SPL~\cite{anderson2018evaluation}, \textbf{Reach SPL} evaluates efficiency of reaching the target room using the reference reach path length $l_{\text{reach},i}$:
\begin{equation}
\text{Reach SPL} = \frac{1}{N} \sum_{i=1}^{N} S_{\text{reach}, i} \cdot \frac{l_{\text{reach}, i}}{\max(p_{\text{reach}, i},\, l_{\text{reach}, i})}.
\end{equation}

\paragraph{REV-SPL (Reach--Explore--Verify SPL).}
Our primary metric \textbf{REV-SPL} extends SPL to capture (i) reaching the target room, (ii) correct \texttt{FOUND}/\texttt{NOT-FOUND} decision, and (iii) sufficient in-room exploration.
Let $S^r_i\!\in\!\{0,1\}$ indicate whether the agent reaches the target room, and let $S^d_i\!\in\!\{0,1\}$ indicate whether it makes the correct final decision.
We define object coverage $C_i\!\in\![0,1]$ as
%\[
%C_i=\frac{\left|\bigcup_{v\in P_i}\texttt{OBJECTSVISIBLE}(v)\right|}{|O_i|},
%\]
$C_i=\frac{\left|\bigcup_{v\in P_i}\texttt{OBJECTSVISIBLE}(v)\right|}{|O_i|},$
where $P_i$ is the agent's in-room trajectory and $O_i$ is the set of annotated object instances in the target room. 
Let $\ell_i$ and $p_i$ denote the reference and actual exploration lengths, respectively.

We compute
\begin{equation}
\begin{aligned}
\mathrm{REV\!-\!SPL}
&=\frac{1}{N}\sum_i S^r_i S^d_i C_i\,
\frac{\ell_i}{\max(p_i,\ell_i)} \\
&\qquad\cdot \min\!\left(1,\frac{p_i}{\ell_i}\right).
\end{aligned}
\label{eq:rev_spl}
\end{equation}

We additionally report \textbf{Coverage} and average path lengths as diagnostics.

\begin{figure*}[t]
    \centering
    \includegraphics[width=\textwidth]{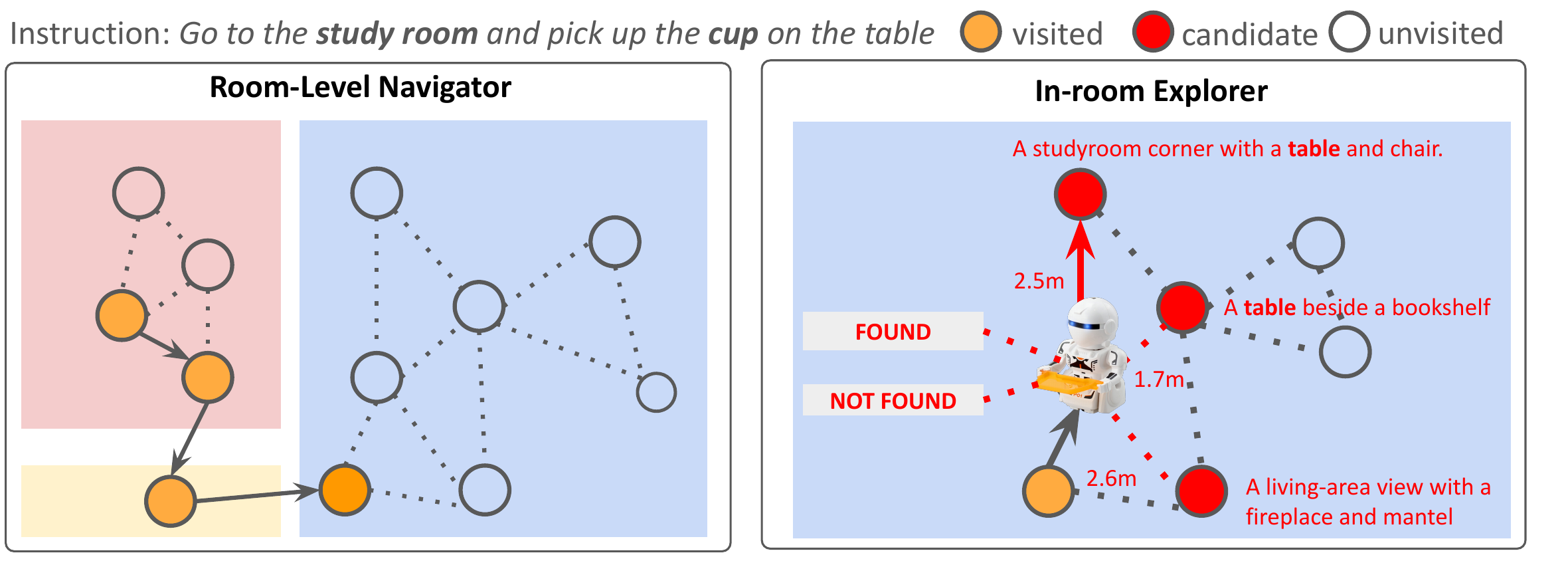}
   \caption{Overview of our two-stage framework, \textbf{\OurModel{}} (Sec.~\ref{sec:room_object_aware_navigation_framework}). 
\textbf{Left} (Sec.~\ref{subsec:coarse-grained}): a room-level navigator trained with \emph{room labels only} guides the agent to enter the target room. 
\textbf{Right} (Sec.~\ref{subsec:fine-grained}): an LLM-based in-room explorer selects the next viewpoint using VLM captions as semantic context and a geometric coverage prior from \textsc{FREE} (Sec.~\ref{subsec:free}) to favor headings that lead to larger unsearched regions.}

    \label{fig:figure3}
\end{figure*} 
\section{\OurModel{} Framework} \label{sec:room_object_aware_navigation_framework}

%%% Figure was thrown to Sec3 for better formatting

%VLN-NF requires (i) reaching the instructed room, (ii) evidence-gathering exploration under partial observability, and (iii) a final \textsc{FOUND}/\texttt{NOT-FOUND} decision. 
%In our pilot study, supervised VLN policies (e.g., DUET) are reliable for inter-room navigation but under-explore after entering the room, while LLM/VLM agents can use semantic priors for in-room search yet struggle with long-horizon navigation. 
%ROAM (Fig.~\ref{fig:figure3}) combines both: a weakly supervised room-level navigator (Sec.~\ref{subsec:coarse-grained}) followed by an LLM/VLM-driven in-room explorer (Sec.~\ref{subsec:fine-grained}), augmented with FREE---a plug-and-play geometric clearance cue for coverage-oriented exploration (Sec.~\ref{subsec:free}).

\OurDataset{} evaluates \emph{evidence-grounded} \texttt{NOT-FOUND} under partial observability: the agent must reach the referenced room, search for the target, and output \texttt{FOUND} or \texttt{NOT-FOUND}. 
This requires reliable room localization and coverage-aware in-room exploration, declaring \texttt{NOT-FOUND} only after sufficient search yields no evidence of the target. 
Note that our goal here is not to formalize absence as a complete belief-theoretic inference problem, but to provide a tractable and reproducible operationalization that can be benchmarked under realistic VLN constraints. 
We observe that supervised VLN policies such as DUET~\cite{chen2022think} are strong at inter-room navigation but often under-explore once inside the target room, since exploration is underdetermined and lacks an explicit ``search finished'' signal. 
In contrast, LLM-based policies~\cite{zhou2024navgpt} can leverage commonsense priors to guide in-room search given context, but they often struggle with floorplan-level navigation under partial observability. 
Motivated by these observations, we propose \OurModelFull{} (\OurModel{}, Fig.~\ref{fig:figure3}), a hybrid framework with a \textbf{Room-Level Navigator} (Sec.~\ref{subsec:coarse-grained}) for robust room localization and an \textbf{In-room Explorer} (Sec.~\ref{subsec:fine-grained}) for evidence-gathering search. 
To compensate for LLMs' weak geometric intuition, we further add a plug-and-play \textbf{Free-space Raycasting Estimation Engine} (\textsc{FREE}) that provides a free-space clearance signal (Sec.~\ref{subsec:free}) to steer exploration toward larger unsearched regions with semantic context.

%\OurDataset{} requires agents to first localize the target room and then explore it to determine whether the task is infeasible. In our pilot study, we observed that supervised models excel at room localization by learning floorplan-level visual patterns, but tend to under-explore the environment. Additionally, collecting VLN instructions requires human annotation, and room-level annotation is generally more affordable. To balance these factors, \OurModelFull{} (\OurModel{}, Figure \ref{fig:figure3}) combines a supervised model for room localization (Section \ref{subsec:coarse-grained})) with an LLM/VLM-based module for room exploration (Section \ref{subsec:fine-grained}). 

\subsection{Room-Level Navigator} \label{subsec:coarse-grained}
We model each environment as a navigation graph \(G=(V,E)\), where each node \(v\in V\) is a panoramic viewpoint and each edge \((u,v)\in E\) denotes a traversable transition.
Let \(r(v)\) be the room label associated with viewpoint \(v\), and let \(r^\star\) be the target room parsed from the instruction.
We define the target-room subgraph as \(R = \{v\in V \mid r(v)=r^\star\}\).

Given an episode starting at \(v_{\text{start}}\), we define an \emph{entry viewpoint}
\begin{equation}
v_{\text{room}} = \arg\min_{v\in R} d_G(v_{\text{start}}, v),
\end{equation}
where \(d_G(\cdot,\cdot)\) is the shortest-path distance on \(G\).
This reduces the original reach-and-explore task to a room-reaching problem: navigate from \(v_{\text{start}}\) to \(v_{\text{room}}\). 
%Importantly, this requires only room labels (no trajectory-level exploration annotation), substantially reducing supervision cost.
In our experiments, we obtain the room-entry target viewpoint from reach-to-room path prefixes (ending at room entry); more generally, since this stage only requires room identity, it can be trained from weak supervision that labels viewpoints with room IDs and synthesizes reach-to-room paths (e.g., shortest paths to any viewpoint in the target room), without trajectory-level \emph{in-room exploration} annotation. 
We adopt DUET~\cite{chen2022think} as the backbone, though any VLN policy capable of room-level navigation can be used.
At inference, the navigator runs until the agent enters the target room (i.e., \(r(v_t)=r^\star\)) or a step budget is reached, after which we hand off to the in-room exploration stage.

\subsection{In-room Explorer} \label{subsec:fine-grained}
After entering the target room \(R\), the In-room Explorer\footnote{Prompts are provided in the Appendix.} searches for the target object \(o'\) and decides whether to output \texttt{FOUND} or \texttt{NOT-FOUND}.
At time \(t\), the agent is at viewpoint \(v_t\in R\) with a history \(H_t\) (visited viewpoints and accumulated evidence), and has a set of reachable neighbors \(\mathcal{N}(v_t)\).

We use a VLM stack to convert observations into semantic context (scene captions and detected objects), and an LLM to select actions in a closed loop.
The action space is
\begin{equation}
\mathcal{A}_t = \{ \texttt{move}(v)\mid v\in\mathcal{N}(v_t) \}\cup \{\texttt{FOUND}, \texttt{NOT-FOUND}\}.
\end{equation}
Let \(D(v,o')\in[0,1]\) denote the confidence from an open-vocabulary detector for object \(o'\) at viewpoint \(v\) (Grounding-DINO~\cite{liu2024grounding} in our implementation).
In addition, for each candidate move \(v\in\mathcal{N}(v_t)\), we compute a free-space clearance cue \(d_{\text{free}}(v_t\!\rightarrow\! v)\) using \textsc{FREE} (Sec.~\ref{subsec:free}).

The explorer terminates with \texttt{FOUND} if \(D(v_t,o')\ge \tau\) at any step; otherwise, it continues exploring and may output \texttt{NOT-FOUND} only when an evidence-backed stopping criterion is satisfied or a hard constraint is met (e.g., frontier exhaustion or a fixed exploration budget).
We implement the explorer by augmenting NavGPT~\cite{zhou2024navgpt} with the \textsc{FREE} clearance cue, where NavGPT tokenizes observations via captioning and open-vocabulary detection and uses an LLM to reason over accumulated evidence and decide whether to continue or terminate.

\subsection{FREE}\label{subsec:free}
When exploring, an agent benefits from both commonsense priors (e.g., cups are likely on tables) and geometric coverage cues (e.g., moving toward a large unsearched area is often informative).
Prior work suggests that LLMs encode commonsense knowledge but can be unreliable at geometric/spatial reasoning~\cite{llmsp1comsa-narayanan-2023-benchmark,llmsp2yan2023inherent,llmsp3cheng2024spatialrgpt,llmsp4yang2024thinking}.
To bridge this gap, we introduce a plug-and-play \textbf{Free-space Raycasting Estimation Engine} (\textsc{FREE}), which provides a \emph{free-space clearance} signal to guide exploration.

\begin{figure}
    \centering
    \includegraphics[width=1\linewidth]{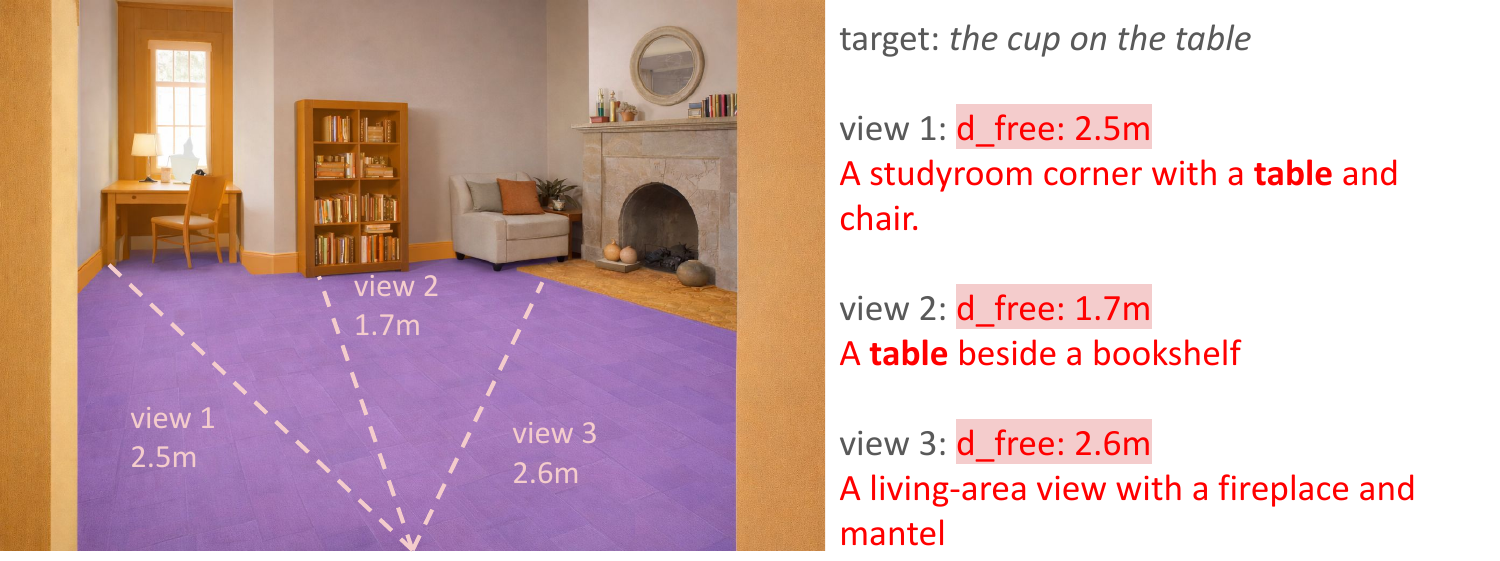}
\caption{\textbf{FREE (Sec.~\ref{subsec:free})} segments navigable floor regions from the current view using an open-vocabulary visual model and uses depth-based raycasting to estimate a free-space clearance \(d_{\text{free}}\) for each candidate direction. The resulting clearance cues are appended to the LLM prompt to encourage coverage-oriented exploration.}    \label{fig:FREE}
\end{figure}

Given the current observation at \(v_t\), \textsc{FREE} estimates, for each candidate heading \(\theta_k\), a clearance distance \(d_{\text{free}}(\theta_k)\) indicating how far the agent can move before encountering an obstacle (Fig.~\ref{fig:FREE}).
Concretely, we segment navigable regions (Grounded-SAM~\cite{ren2024grounded} in our experiments) and back-project the mask into 3D to perform raycasting.
We then map each candidate move \(v\in\mathcal{N}(v_t)\) to its heading \(\theta(v)\) and attach \(d_{\text{free}}(\theta(v))\) to the planner prompt as a geometric prior, encouraging actions that lead to larger unsearched regions.

\section{Experiments}

%%%% Table 1 was thrown to sec 3 for better formatting $$$$
\begin{table*}[t]
\centering
\resizebox{\textwidth}{!}{%
\begin{tabular}{l|l|rrr|cc|cc}
\hline
Setting & Method & Coverage & Len (reach) & Len (explore) & Reach SR & Reach\&Decision SR & Reach SPL & \OurMetric{} \\
\hline
supervised   & DUET                      & 69.5\% & 13.1 & 7.8  & 53.8\% & 33.8\% & 37.0\% & 4.2\% \\
unsupervised & NavGPT                    & 59.1\% & 10.6 & 2.7 & 8.2\%  & 5.4\%  & 7.0\% & 1.0\% \\
unsupervised & MapGPT                    & 63.3\% & 13.3 & 3.5 & 30.0\%  & 14.0\%  & 18.2\% & 3.2\% \\
unsupervised & SoM-Gemini-2.0-Flash & 80.9\% & 10.0 & 14.2 & 39.4\% & 22.0\% & 28.9\% & 1.5\% \\
hybrid       & \OurModel{} (Ours)-GPT-3.5       & 82.1\% & 11.2 & 9.9  & 58.6\% & 37.6\% & 44.1\% & \textbf{6.1\%} \\
hybrid       & \OurModel{} (Ours)-GPT-4o        & \textbf{82.8}\% & 12.8 & 22.0 & \textbf{62.6}\% & \textbf{41.4}\% & \textbf{45.4\%} & 5.6\% \\
\hline
\end{tabular}
}
\caption{
Comparison with baselines on \OurDataset{} \texttt{val\_unseen}. We report our primary metric \textbf{REV-SPL} (Eq.~\ref{eq:rev_spl}), together with decomposed diagnostics, \textbf{Reach SR}, \textbf{Reach\&Decision SR}, and \textbf{Reach SPL}.
 (Sec. \ref{subsec:results})}
\label{table:baseline}
\end{table*}

\subsection{Baseline Methods}

We adapt several representative VLN baselines to our dataset \OurDataset{}.
\textbf{DUET}~\cite{chen2022think} is extended by adding an additional \texttt{NOT-FOUND} class in its decision head to support abstention on infeasible episodes.
\textbf{NavGPT}~\cite{zhou2024navgpt} is modified to include an explicit \texttt{NOT-FOUND} action in its action space/prompt, allowing the agent to terminate with \texttt{NOT-FOUND} when appropriate. 
\textbf{MapGPT}~\cite{chen2024mapgpt} is a map-guided GPT-based VLN agent that constructs an online linguistic topological map and performs adaptive multi-step path planning. 
\textbf{SoM-Gemini-2.0-Flash} follows the Set-of-Marks prompting strategy~\cite{yang2023set} and uses Gemini-2.0-Flash as the multimodal backbone, providing a captioning-free alternative to NavGPT’s captioning-then-LLM pipeline.
Concretely, we overlay candidate viewpoints in Matterport3D with red markers and numeric labels, and query Gemini to select the next action (navigate, stop/\texttt{FOUND}, or \texttt{NOT-FOUND}). 
We include it as a planner/map-augmented baseline and adapt it to \OurDataset{} by evaluating it under the same \texttt{FOUND}/\texttt{NOT-FOUND} decision protocol and evaluation pipeline as the other methods.

%\textbf{Gemini} To evaluate the performance of a Multimodal LLM on our task, we built a simple baseline using Gemini 2.0 Flash \cite{google2024gemini2flashdocs}. We adopt a Set-of-Mark approach\cite{yang2023set}, where we annotate the candidate viewpoints with red dots and numerical labels directly on the observations at each viewpoint in the Matterport3D simulator. These annotated images are then fed into Gemini, which is tasked with selecting the next action—whether to move to a specific viewpoint, stop the navigation, or indicate that a target object has been found.

\subsection{Results}\label{subsec:results}
As shown in Table~\ref{table:baseline}, \OurModel{} achieves the best performance among all compared baselines on \OurDataset{} under our primary metric \OurMetric{}.
\paragraph{Findings.}
\noindent\textbf{(1) Room reaching vs.\ in-room verification.} DUET achieves strong room-reaching (Reach SR 53.8\%, Reach SPL 37.0\%) but attains only 69.5\% coverage and 4.2 REV-SPL, indicating that reaching the target room does not guarantee evidence-gathering verification. 
\textbf{(2) LLM/VLM baselines are unreliable under partial observability.} NavGPT rarely reaches the target room (Reach SR 8.2\%) and yields 1.0 REV-SPL; while SoM explores more (Coverage 80.9\%, Len(explore) 14.2), it still falls short on joint reach+decision (Reach\&Decision SR 22.0\%, REV-SPL 1.5), showing that commonsense alone is insufficient for robust long-horizon navigation and stopping. 
MapGPT improves over other LLM-based baselines (REV-SPL 3.2 vs.\ 1.0 for NavGPT and 1.5 for SoM), indicating that stronger global planning helps under false-premise instructions.
However, it still remains below DUET (4.2) and both ROAM variants (6.1/5.6), suggesting that map-guided planning alone is insufficient for evidence-grounded \texttt{FOUND}/\texttt{NOT-FOUND} decisions under partial observability. 
\textbf{(3) ROAM closes both gaps.} By combining weakly supervised navigation with LLM/VLM-driven in-room search, ROAM substantially improves both reaching and decision quality (Reach SR 58.6/62.6\%, Reach\&Decision SR 37.6/41.4\%) and achieves the best REV-SPL (6.1/5.6), outperforming DUET by 45\% while using only room-level supervision. 
\textbf{(4) REV-SPL captures the efficiency--thoroughness trade-off.} ROAM-GPT-4o attains higher Reach\&Decision SR than ROAM-GPT-3.5 (41.4\% vs.\ 37.6\%) but explores much longer (22.0 vs.\ 9.9), leading to slightly lower REV-SPL (5.6 vs.\ 6.1), consistent with REV-SPL penalizing inefficient wandering despite correct decisions.

%Table \ref{table:baseline} compares \OurModel{} with baseline agents. 
%\OurModel{} attains the highest \textit{Reach\&Found SR} and \textit{Reach\&Found SPL} among all models. 
%DUET performs poorly: we find its exploration is limited (as reflected by its low object coverage rate) and it exhibits early stopping (very short exploration length), resulting in low Reach\&Found SR and reduced practical usefulness, as it cannot efficiently handle infeasible instructions in real-world scenarios.
%LLM-only approaches, NavGPT and Gemini, struggle to reach the room due to limitations in visual perception and the lack of spatial understanding in current LLM/VLM pretraining, resulting in low overall performance.

%In contrast, our \OurModel{} achieves leading performance across metrics. 
%The Coarse-grained Stage guides the agent into the correct room, improving Reach SR and SPL, while the Fine-grained Stage enhances reasoning for accurate Found/Not Found decisions. 
%This staged design, requiring only room-level supervision, effectively combines coarse navigation with fine-level object discovery.

\paragraph{Ablation Study.}
%\textbf{Ablation Study (Table~\ref{table:ablation}) 
\textbf{Table~\ref{table:ablation} demonstrates the effectiveness of our two-stage hybrid framework and the \textsc{FREE} design.}
 Comparing (1) and (2), adding a room-level navigator trained with \emph{room labels only} dramatically improves performance (e.g., Reach SR 7.8\%$\rightarrow$59.4\% and \OurMetric{} 0.8\%$\rightarrow$5.6\%), indicating that an LLM-only controller struggles with long-horizon room reaching under partial observability. 
Comparing (2) and (3), \textsc{FREE} further increases evidence coverage (79.2\%$\rightarrow$\textbf{82.1\%}) and improves \OurMetric{} (5.6\%$\rightarrow$\textbf{6.1\%}), validating that geometric clearance cues complement caption-based commonsense by steering exploration toward larger unsearched regions, while leaving reaching performance essentially unchanged.

\paragraph{Transfer to standard feasible VLN.}
\textbf{(Table~\ref{table:ablationREVERIE}) shows that \OurModel{} also enhances VLN performance on REVERIE \texttt{val\_unseen}.}
Removing the room-level navigator again leads to very poor navigation (Reach SR 8.1\%, SPL 6.8\%), while the two-stage design substantially improves both reaching and task success (Reach SR 62.0\%, SPL 23.7\%). 
Importantly, \textbf{\textsc{FREE} remains beneficial} even in the feasible-only setting, improving SR (42.0\%$\rightarrow$45.2\%) and SPL (23.7\%$\rightarrow$25.1\%), suggesting that FREE-style coverage cues help the LLM planner choose more informative viewpoints without hurting conventional VLN performance.

\begin{table}
\centering

\resizebox{\columnwidth}{!}{%

    \begin{tabular}{l|ll|lllll}
    \hline
    \# & 2 Stage & FREE &  Coverage  & Reach SR & R\&D SR & Reach SPL & \OurMetric{} \\\hline
        (1) & \XSolidBrush & \XSolidBrush  & 75.7\% & 7.8\% & 4.4\% & 6.7\% & 0.8\% \\
        (2) & \Checkmark & \XSolidBrush &  79.2\% & \textbf{59.4\%} & 37.2\% & \textbf{44.2\%} & 5.6\%  \\
        (3) & \Checkmark & \Checkmark & \textbf{82.1}\% & 58.6\% & \textbf{37.6\%} & 44.1\% & \textbf{6.1\%} \\
    \hline
    \end{tabular}
}
\caption{\label{table:ablation} Ablation study of our ROAM framework using GPT-3.5 on the val\_unseen split of \OurDataset{}.}
\end{table}

\begin{table}
\centering

\resizebox{\columnwidth}{!}{%

    \begin{tabular}{l|ll|llll}
    \hline
    \# & 2 Stage & FREE &  Reach SR  & Reach SPL  & SR &  SPL \\\hline
        (1) & \XSolidBrush & \XSolidBrush & 8.1\% & 6.6\% & 9.2\% &  6.8\% \\
        (2) & \Checkmark & \XSolidBrush &  62.0\% & 48.1\% & 42.0\% & 23.7\%  \\
        (3) & \Checkmark & \Checkmark & 64.0\% & 48.4\% & 45.2\% &  25.1\%  \\
    \hline
    \end{tabular}
}
\caption{\label{table:ablationREVERIE}ROAM (GPT-3.5) on REVERIE val\_unseen.}
\end{table}

\paragraph{False \texttt{NOT-FOUND} on feasible episodes.}
As shown in Table \ref{tab:false_nf_found}, on \OurDataset{} val\_unseen, we further audit decision-level false \texttt{NOT-FOUND} predictions made by ROAM (GPT-3.5) on \texttt{FOUND} episodes. Most such errors are classic room-reaching failures (55.7\%), while the remainder are dominated by perception/grounding uncertainty (31.0\%) and a smaller exploration-control component (13.3\%), suggesting that enabling \texttt{NOT-FOUND} mainly amplifies calibration and verification challenges rather than introducing a fundamentally new failure mode.

\begin{table}[t]
\centering
\small
\setlength{\tabcolsep}{5pt}
\begin{tabular}{lc}
\toprule
\textbf{Error source} & \textbf{Share} \\
\midrule
Room-reaching failure & 55.7\% \\
Perception/grounding uncertainty & 31.0\% \\
Exploration-control failure & 13.3\% \\
\bottomrule
\end{tabular}
\caption{
Error-source decomposition of decision-level false \texttt{NOT-FOUND} predictions made by ROAM (GPT-3.5) on feasible (\texttt{FOUND}) episodes in \OurDataset{} val\_unseen.
}
\label{tab:false_nf_found}
\end{table}
\section{Conclusion}
We study \emph{evidence-grounded} \texttt{NOT-FOUND} for 3D VLN under unreliable (false-premise) instructions.
We introduce \OurDataset{}, constructed by pairing feasible REVERIE episodes with automatically rewritten-and-verified infeasible counterparts, and propose \OurMetric{} to evaluate reaching, decision correctness, and evidence-gathering exploration efficiency.
We further present ROAM, a two-stage hybrid that combines room-level supervision for robust room localization with VLM/LLM-driven in-room exploration and verification, augmented by a free-space clearance cue (\textsc{FREE}).
Across baselines, ROAM achieves the best performance on \OurDataset{} (6.1 REV-SPL), improving over a strong supervised agent by 45\% while requiring only room-level annotations. We view ROAM as a structured first step toward evidence-grounded abstention in 3D VLN under partial observability, while more principled absence reasoning and broader recovery behaviors remain important future directions.

%We propose a new benchmark, \OurDataset{}, extending VLN to real-world scenarios with potentially erroneous human instructions leading to not-found targets. 
%Our pipeline uses LLMs to generate such instructions and a grounding model to verify object absence, simulating unreliable guidance. 

%We introduce metrics to evaluate agents' ability to reach target rooms, explore effectively, and issue \texttt{NOT-FOUND} signals. 
%Experiments show existing VLN models struggle in this setting. 
%To address this, we propose ROAM, a two-stage framework that improves performance while highlighting the need for more robust, error-aware navigation methods.
%\newpage

%\textbf{Data coverage}. The experiments in this paper are based on the REVERIE dataset, which we adapted to suit the \OurDataset{} setting. To ensure the validity of automatically generated paths during the path generation process, we filtered out instructions that are incompatible with the \OurDataset{} task. 
%As a result, only around 45\% of the original instructions were retained, potentially limiting the diversity and coverage of the final dataset.
%To ensure reliable verification and reference exploration generation, we apply strict filtering; as a result, VLN-NF currently retains ~45\% of REVERIE episodes. This trades coverage for higher confidence in NF correctness and evaluation stability.
\section*{Limitations}
Our study has several limitations that point to directions for future work.

\paragraph{Data coverage and filtering.}
The experiments in this paper are based on REVERIE, which we adapt to the \OurDataset{} setting.
To ensure reliable verification and stable reference-exploration generation, we apply strict filtering and currently retain 45\% of REVERIE episodes.
This trades raw coverage for higher confidence in NF correctness and evaluation stability.
Importantly, this filtering is intended as a quality-control gate for evaluability rather than a mechanism to skew the benchmark distribution.
As shown in Appendix, the filtering-only subset induces only very small language/object distribution shift relative to full REVERIE, while the additional shift in final \OurDataset{} is modest and expected from substituting absent-but-plausible targets.

\textbf{Evaluation of path quality}. While we conducted human evaluation on a sampled subset of the \OurDataset{} dataset to confirm that the automatically labeled \texttt{NOT-FOUND} targets are accurate—achieving approximately 98\% correctness—we did not perform a formal assessment of the quality or naturalness of the generated paths themselves. This omission may leave potential gaps in evaluating whether the paths are contextually and semantically appropriate.

\textbf{Grounding-confidence threshold sensitivity in ROAM}. The performance of the ROAM framework is sensitive to the object grounding model’s (e.g., Grounding-DINO) confidence threshold. Since the \texttt{NOT-FOUND} signal in our system is triggered based on whether any object surpasses the detection threshold, this fixed threshold may lead to performance degradation when applied to environments that differ significantly from the original training domain.

\paragraph{Scope of false-premise types.}
VLN-NF currently instantiates a focused form of goal-level false premise: the referenced target object is absent from the room specified by the instruction.
We choose this setting as a controlled but practical starting point, since mis-specified object locations are common in human instructions and, under partial observability, absence cannot be verified from a single viewpoint.
Importantly, several broader false-premise variants reduce to the same core decision problem once conditioned on a described region; for example, a target located in the wrong room corresponds to ``absent in the specified room,''
and wrong attribute/object descriptions correspond to ``absence of the described entity.''
However, cases with multiple plausible referents or under-specified instructions may require clarification or interactive recovery rather than pure absence verification.
Extending VLN-NF to cover these broader variants and richer recovery behaviors is an important direction for future work.

\textbf{Future Work: Recovery Policy.} Once the system identifies an instruction as leading to a \texttt{NOT-FOUND} target, it currently terminates navigation. However, in real-world scenarios, users may expect recovery behaviors, such as rephrasing the instruction, exploring alternative paths, or asking for clarification. Designing and integrating such recovery strategies remains an open area for future research.

\paragraph{Acknowledgment}
This work was supported in part by National Science and Technology Council, Taiwan, under Grant NSTC 113-2634-F-002-007.

\bibliography{custom}

@inproceedings{burns2022motif,
  author       = {Andrea Burns and Deniz Arsan and Sanjna Agrawal and Ranjitha Kumar and Kate Saenko and Bryan A. Plummer},
  editor       = {Shai Avidan and Gabriel J. Brostow and Moustapha Ciss{\'{e}} and Giovanni Maria Farinella and Tal Hassner},
  title        = {A Dataset for Interactive Vision-Language Navigation with Unknown Command Feasibility},
  booktitle    = {Computer Vision -- ECCV 2022 - 17th European Conference, Tel Aviv, Israel, October 23--27, 2022, Proceedings, Part VIII},
  series       = {Lecture Notes in Computer Science},
  volume       = {13668},
  pages        = {312--328},
  publisher    = {Springer},
  year         = {2022},
  doi          = {10.1007/978-3-031-20074-8_18},
  url          = {https://doi.org/10.1007/978-3-031-20074-8_18}
}

@inproceedings{taioli2024mind,
  author    = {Francesco Taioli and Stefano Rosa and Alberto Castellini and Lorenzo Natale and Alessio Del Bue and Alessandro Farinelli and Marco Cristani and Yiming Wang},
  title     = {Mind the Error! Detection and Localization of Instruction Errors in Vision-and-Language Navigation},
  booktitle = {2024 IEEE/RSJ International Conference on Intelligent Robots and Systems (IROS)},
  year      = {2024},
  pages     = {12993--13000},
  doi       = {10.1109/IROS58592.2024.10801822},
  url       = {https://ieeexplore.ieee.org/document/10801822/}
}

@article{hong2024r2runo,
  title   = {Navigating Beyond Instructions: Vision-and-Language Navigation in Obstructed Environments},
  author  = {Haodong Hong and Sen Wang and Zi Huang and Qi Wu and Jiajun Liu},
  journal = {arXiv preprint arXiv:2407.21452},
  year    = {2024},
  doi     = {10.48550/arXiv.2407.21452},
  note    = {Accepted to ACM Multimedia (MM) 2024}
}

@inproceedings{hsieh2025dowhat,
  title     = {Do What? Teaching Vision-Language-Action Models to Reject the Impossible},
  author    = {Wen-Han Hsieh and Elvis Hsieh and Dantong Niu and Trevor Darrell and Roei Herzig and David M. Chan},
  booktitle = {Findings of the Association for Computational Linguistics: EMNLP 2025},
  year      = {2025},
  month     = nov,
  address   = {Suzhou, China},
  publisher = {Association for Computational Linguistics},
  pages     = {11861--11869},
  doi       = {10.18653/v1/2025.findings-emnlp.635},
  url       = {https://aclanthology.org/2025.findings-emnlp.635/}
}

@inproceedings{chakraborty2025heal,
  title     = {{HEAL}: An Empirical Study on Hallucinations in Embodied Agents Driven by Large Language Models},
  author    = {Trishna Chakraborty and Udita Ghosh and Xiaopan Zhang and Fahim Faisal Niloy and Yue Dong and Jiachen Li and Amit Roy-Chowdhury and Chengyu Song},
  booktitle = {Findings of the Association for Computational Linguistics: EMNLP 2025},
  year      = {2025},
  month     = nov,
  address   = {Suzhou, China},
  publisher = {Association for Computational Linguistics},
  pages     = {21226--21243},
  doi       = {10.18653/v1/2025.findings-emnlp.1158},
  url       = {https://aclanthology.org/2025.findings-emnlp.1158/}
}

@article{wu2024noisyeqa,
  title   = {NoisyEQA: Benchmarking Embodied Question Answering Against Noisy Queries},
  author  = {Tao Wu and Chuhao Zhou and Yen Heng Wong and Lin Gu and Jianfei Yang},
  journal = {arXiv preprint arXiv:2412.10726},
  year    = {2024},
  doi     = {10.48550/arXiv.2412.10726},
  url     = {https://arxiv.org/abs/2412.10726}
}

@article{wu2025abstaineqa,
  title   = {When Robots Should Say ``I Don't Know'': Benchmarking Abstention in Embodied Question Answering},
  author  = {Tao Wu and Chuhao Zhou and Guangyu Zhao and Haozhi Cao and Yewen Pu and Jianfei Yang},
  journal = {arXiv preprint arXiv:2512.04597},
  year    = {2025},
  doi     = {10.48550/arXiv.2512.04597},
  url     = {https://arxiv.org/abs/2512.04597}
}

@InProceedings{Kamath_2023_CVPR,
  author    = {Kamath, Aishwarya and Anderson, Peter and Wang, Su and Koh, Jing Yu and Ku, Alexander and Waters, Austin and Yang, Yinfei and Baldridge, Jason and Parekh, Zarana},
  title     = {A New Path: Scaling Vision-and-Language Navigation With Synthetic Instructions and Imitation Learning},
  booktitle = {Proceedings of the IEEE/CVF Conference on Computer Vision and Pattern Recognition (CVPR)},
  month     = {June},
  year      = {2023},
  pages     = {10813--10823},
  url       = {https://openaccess.thecvf.com/content/CVPR2023/html/Kamath_A_New_Path_Scaling_Vision-and-Language_Navigation_With_Synthetic_Instructions_and_CVPR_2023_paper.html}
}

@inproceedings{seo2024drildice,
  title     = {Mitigating Covariate Shift in Behavioral Cloning via Robust Stationary Distribution Correction},
  author    = {Seo, Seokin and Lee, Byung-Jun and Lee, Jongmin and Hwang, HyeongJoo and Yang, Hongseok and Kim, Kee-Eung},
  booktitle = {Advances in Neural Information Processing Systems},
  year      = {2024},
  volume    = {37},
  url       = {https://proceedings.neurips.cc/paper_files/paper/2024/hash/c556da88a2665e6266453d8c9b8a552d-Abstract-Conference.html}
}

@inproceedings{zhang2024diffusionmeetsdagger,
  title     = {Diffusion Meets {DAgger}: Supercharging Eye-in-hand Imitation Learning},
  author    = {Zhang, Xiaoyu and Chang, Matthew and Kumar, Pranav and Gupta, Saurabh},
  booktitle = {Robotics: Science and Systems (RSS)},
  year      = {2024},
  eprint    = {2402.17768},
  archivePrefix = {arXiv},
  primaryClass  = {cs.RO},
  doi       = {10.48550/arXiv.2402.17768},
  url       = {https://arxiv.org/abs/2402.17768}
}

@inproceedings{xiang-etal-2020-learning,
  title     = {Learning to Stop: A Simple yet Effective Approach to Urban Vision-Language Navigation},
  author    = {Xiang, Jiannan and Wang, Xin and Wang, William Yang},
  editor    = {Cohn, Trevor and He, Yulan and Liu, Yang},
  booktitle = {Findings of the Association for Computational Linguistics: EMNLP 2020},
  month     = nov,
  year      = {2020},
  address   = {Online},
  publisher = {Association for Computational Linguistics},
  url       = {https://aclanthology.org/2020.findings-emnlp.62/},
  doi       = {10.18653/v1/2020.findings-emnlp.62},
  pages     = {699--707}
}

@inproceedings{
su2024contextaware,
title={Context-Aware Replanning with Pre-Explored Semantic Map for Object Navigation},
author={Po-Chen Ko and Hung-Ting Su and CY Chen and Jia-Fong Yeh and Min Sun and Winston H. Hsu},
booktitle={8th Annual Conference on Robot Learning},
year={2024},
url={https://openreview.net/forum?id=Dftu4r5jHe}
}

@misc{chen2026adaptbenchmarkingcommonsenseplanning,
title={ADAPT: Benchmarking Commonsense Planning under Unspecified Affordance Constraints},
author={Pei-An Chen and Yong-Ching Liang and Jia-Fong Yeh and Hung-Ting Su and Yi-Ting Chen and Min Sun and Winston Hsu},
year={2026},
eprint={2604.14902},
archivePrefix={arXiv},
primaryClass={cs.AI},
url={https://arxiv.org/abs/2604.14902},
}

@inproceedings{llmsp1comsa-narayanan-2023-benchmark,
  title     = {A Benchmark for Reasoning with Spatial Prepositions},
  author    = {Comsa, Iulia and Narayanan, Srini},
  booktitle = {Proceedings of the 2023 Conference on Empirical Methods in Natural Language Processing},
  pages     = {16328--16335},
  address   = {Singapore},
  publisher = {Association for Computational Linguistics},
  year      = {2023},
  doi       = {10.18653/v1/2023.emnlp-main.1015},
  url       = {https://aclanthology.org/2023.emnlp-main.1015/}
}

@misc{llmsp2yan2023inherent,
  title         = {Inherent Limitations of GPT=4 Regarding Spatial Information},
  author        = {He Yan and Xinyao Hu and Xiangpeng Wan and Chengyu Huang and Kai Zou and Shiqi Xu},
  year          = {2023},
  eprint        = {2312.03042},
  archivePrefix = {arXiv},
  primaryClass  = {cs.CL},
  url           = {https://arxiv.org/abs/2312.03042}
}

@inproceedings{llmsp3cheng2024spatialrgpt,
  title     = {SpatialRGPT: Grounded Spatial Reasoning in Vision-Language Models},
  author    = {Cheng, An-Chieh and Yin, Hongxu and Fu, Yang and Guo, Qiushan and Yang, Ruihan and Kautz, Jan and Wang, Xiaolong and Liu, Sifei},
  booktitle = {Advances in Neural Information Processing Systems},
  year      = {2024}
}

@misc{llmsp4yang2024thinking,
  title         = {Thinking in Space: How Multimodal Large Language Models See, Remember, and Recall Spaces},
  author        = {Jihan Yang and Shusheng Yang and Anjali W. Gupta and Rilyn Han and Li Fei-Fei and Saining Xie},
  year          = {2024},
  eprint        = {2412.14171},
  archivePrefix = {arXiv},
  primaryClass  = {cs.CV},
  url           = {https://arxiv.org/abs/2412.14171}
}

@inproceedings{anderson2018vision,
  title={Vision-and-language navigation: Interpreting visually-grounded navigation instructions in real environments},
  author={Anderson, Peter and Wu, Qi and Teney, Damien and Bruce, Jake and Johnson, Mark and S{\"u}nderhauf, Niko and Reid, Ian and Gould, Stephen and Van Den Hengel, Anton},
  booktitle={Proceedings of the IEEE conference on computer vision and pattern recognition},
  pages={3674--3683},
  year={2018}
}

@inproceedings{qi2020reverie,
  title={Reverie: Remote embodied visual referring expression in real indoor environments},
  author={Qi, Yuankai and Wu, Qi and Anderson, Peter and Wang, Xin and Wang, William Yang and Shen, Chunhua and Hengel, Anton van den},
  booktitle={Proceedings of the IEEE/CVF Conference on Computer Vision and Pattern Recognition},
  pages={9982--9991},
  year={2020}
}

@article{chang2017matterport3d,
  title={Matterport3d: Learning from rgb-d data in indoor environments},
  author={Chang, Angel and Dai, Angela and Funkhouser, Thomas and Halber, Maciej and Niessner, Matthias and Savva, Manolis and Song, Shuran and Zeng, Andy and Zhang, Yinda},
  journal={arXiv preprint arXiv:1709.06158},
  year={2017}
}

@article{ku2020room,
  title={Room-across-room: Multilingual vision-and-language navigation with dense spatiotemporal grounding},
  author={Ku, Alexander and Anderson, Peter and Patel, Roma and Ie, Eugene and Baldridge, Jason},
  journal={arXiv preprint arXiv:2010.07954},
  year={2020}
}

@inproceedings{thomason2020vision,
  title={Vision-and-dialog navigation},
  author={Thomason, Jesse and Murray, Michael and Cakmak, Maya and Zettlemoyer, Luke},
  booktitle={Conference on Robot Learning},
  pages={394--406},
  year={2020},
  organization={PMLR}
}

@inproceedings{pashevich2021episodic,
  title={Episodic transformer for vision-and-language navigation},
  author={Pashevich, Alexander and Schmid, Cordelia and Sun, Chen},
  booktitle={Proceedings of the IEEE/CVF International Conference on Computer Vision},
  pages={15942--15952},
  year={2021}
}

@article{chen2021history,
  title={History aware multimodal transformer for vision-and-language navigation},
  author={Chen, Shizhe and Guhur, Pierre-Louis and Schmid, Cordelia and Laptev, Ivan},
  journal={Advances in neural information processing systems},
  volume={34},
  pages={5834--5847},
  year={2021}
}

@inproceedings{chen2022think,
  title={Think global, act local: Dual-scale graph transformer for vision-and-language navigation},
  author={Chen, Shizhe and Guhur, Pierre-Louis and Tapaswi, Makarand and Schmid, Cordelia and Laptev, Ivan},
  booktitle={Proceedings of the IEEE/CVF Conference on Computer Vision and Pattern Recognition},
  pages={16537--16547},
  year={2022}
}

@inproceedings{wang2023gridmm,
  title={Gridmm: Grid memory map for vision-and-language navigation},
  author={Wang, Zihan and Li, Xiangyang and Yang, Jiahao and Liu, Yeqi and Jiang, Shuqiang},
  booktitle={Proceedings of the IEEE/CVF International conference on computer vision},
  pages={15625--15636},
  year={2023}
}

@inproceedings{georgakis2022cross,
  title={Cross-modal map learning for vision and language navigation},
  author={Georgakis, Georgios and Schmeckpeper, Karl and Wanchoo, Karan and Dan, Soham and Miltsakaki, Eleni and Roth, Dan and Daniilidis, Kostas},
  booktitle={Proceedings of the IEEE/CVF conference on computer vision and pattern recognition},
  pages={15460--15470},
  year={2022}
}

@inproceedings{zhou2024navgpt,
  title={Navgpt: Explicit reasoning in vision-and-language navigation with large language models},
  author={Zhou, Gengze and Hong, Yicong and Wu, Qi},
  booktitle={Proceedings of the AAAI Conference on Artificial Intelligence},
  volume={38},
  number={7},
  pages={7641--7649},
  year={2024}
}

@inproceedings{zhou2024navgpt2,
  title={Navgpt-2: Unleashing navigational reasoning capability for large vision-language models},
  author={Zhou, Gengze and Hong, Yicong and Wang, Zun and Wang, Xin Eric and Wu, Qi},
  booktitle={European Conference on Computer Vision},
  pages={260--278},
  year={2024},
  organization={Springer}
}

@article{pan2023langnav,
  title={Langnav: Language as a perceptual representation for navigation},
  author={Pan, Bowen and Panda, Rameswar and Jin, SouYoung and Feris, Rogerio and Oliva, Aude and Isola, Phillip and Kim, Yoon},
  journal={arXiv preprint arXiv:2310.07889},
  year={2023}
}

@inproceedings{qiao2024llm,
  title={LLM as Copilot for Coarse-Grained Vision-and-Language Navigation},
  author={Qiao, Yanyuan and Liu, Qianyi and Liu, Jiajun and Liu, Jing and Wu, Qi},
  booktitle={European Conference on Computer Vision},
  pages={459--476},
  year={2024},
  organization={Springer}
}

@inproceedings{chen2024mapgpt,
  title={MapGPT: Map-Guided Prompting with Adaptive Path Planning for Vision-and-Language Navigation},
  author={Chen, Jiaqi and Lin, Bingqian and Xu, Ran and Chai, Zhenhua and Liang, Xiaodan and Wong, Kwan-Yee~K.},
  booktitle = "Proceedings of the 62nd Annual Meeting of the Association for Computational Linguistics",
  year={2024}
}

@article{lin2025navcot,
  title={Navcot: Boosting llm-based vision-and-language navigation via learning disentangled reasoning},
  author={Lin, Bingqian and Nie, Yunshuang and Wei, Ziming and Chen, Jiaqi and Ma, Shikui and Han, Jianhua and Xu, Hang and Chang, Xiaojun and Liang, Xiaodan},
  journal={IEEE Transactions on Pattern Analysis and Machine Intelligence},
  year={2025},
  publisher={IEEE}
}

@inproceedings{long2024discuss,
  title={Discuss before Moving: Visual Language Navigation via Multi-Expert Discussions},
  author={Long, Yuxing and Li, Xiaoqi and Cai, Wenzhe and Dong, Hao},
  booktitle={Proceedings of the IEEE International Conference on Robotics and Automation (ICRA)},
  year={2024},
  organization={IEEE}
}

@inproceedings{hahn2023way,
  author = {Meera Hahn and Amit Raj and James M. Rehg},
  title = {Which way is ‘right’?: Uncovering limitations of Vision-and-Language Navigation models},
  booktitle = {Proceedings of the International Conference on Autonomous Agents and Multiagent Systems (AAMAS)},
  year = {2023},
  address = {Richland, SC},
  pages = {2415--2417},
  url = {https://api.semanticscholar.org/CorpusID:253381693}
}

@inproceedings{wang2002study,
  title={A Study of Object-Location Memory},
  author={Wang, Hongbin and Johnson, Todd R and Zhang, Jiajie and Wang, Yue},
  booktitle={Proceedings of the Annual Meeting of the Cognitive Science Society},
  volume={24},
  year={2002},
  url={https://escholarship.org/uc/item/62s8d5p0}
}

@inproceedings{li2022grounded,
  title={Grounded language-image pre-training},
  author={Li, Liunian Harold and Zhang, Pengchuan and Zhang, Haotian and Yang, Jianwei and Li, Chunyuan and Zhong, Yiwu and Wang, Lijuan and Yuan, Lu and Zhang, Lei and Hwang, Jenq-Neng and others},
  booktitle={Proceedings of the IEEE/CVF conference on computer vision and pattern recognition},
  pages={10965--10975},
  year={2022}
}

@article{anderson2018evaluation,
  title={On evaluation of embodied navigation agents},
  author={Anderson, Peter and Chang, Angel and Chaplot, Devendra Singh and Dosovitskiy, Alexey and Gupta, Saurabh and Koltun, Vladlen and Kosecka, Jana and Malik, Jitendra and Mottaghi, Roozbeh and Savva, Manolis and others},
  journal={arXiv preprint arXiv:1807.06757},
  year={2018}
}

@article{ren2024grounded,
  title={Grounded sam: Assembling open-world models for diverse visual tasks},
  author={Ren, Tianhe and Liu, Shilong and Zeng, Ailing and Lin, Jing and Li, Kunchang and Cao, He and Chen, Jiayu and Huang, Xinyu and Chen, Yukang and Yan, Feng and others},
  journal={arXiv preprint arXiv:2401.14159},
  year={2024}
}

@inproceedings{liu2024grounding,
  title={Grounding dino: Marrying dino with grounded pre-training for open-set object detection},
  author={Liu, Shilong and Zeng, Zhaoyang and Ren, Tianhe and Li, Feng and Zhang, Hao and Yang, Jie and Jiang, Qing and Li, Chunyuan and Yang, Jianwei and Su, Hang and others},
  booktitle={European Conference on Computer Vision},
  pages={38--55},
  year={2024},
  organization={Springer}
}

@misc{openai2023gpt4,
  author = {OpenAI},
  title = {GPT-4 Technical Report},
  year = {2023},
  url = {https://openai.com/research/gpt-4},
  note = {Accessed: 2025-05-18}
}

@article{yang2023set,
  title={Set-of-mark prompting unleashes extraordinary visual grounding in gpt-4v},
  author={Yang, Jianwei and Zhang, Hao and Li, Feng and Zou, Xueyan and Li, Chunyuan and Gao, Jianfeng},
  journal={arXiv preprint arXiv:2310.11441},
  year={2023}
}

@article{georgiev2024gemini,
  title={Gemini 1.5: Unlocking multimodal understanding across millions of tokens of context},
  author={Georgiev, Petko and Lei, Ving Ian and Burnell, Ryan and Bai, Libin and Gulati, Anmol and Tanzer, Garrett and Vincent, Damien and Pan, Zhufeng and Wang, Shibo and Mariooryad, Soroosh and others},
  journal={arXiv preprint arXiv:2403.05530},
  year={2024},
  url={https://arxiv.org/abs/2403.05530}
}

@inproceedings{zhang2024navhint,
  title={Navhint: Vision and language navigation agent with a hint generator},
  author={Zhang, Yue and Guo, Quan and Kordjamshidi, Parisa},
  year={2024},
  organization={Association for Computational Linguistics}
}

@inproceedings{krantzvlnce2020,
  title={Beyond the Nav-Graph: Vision and Language Navigation in Continuous Environments},
  author={Jacob Krantz and Erik Wijmans and Arjun Majundar and Dhruv Batra and Stefan Lee},
  booktitle={European Conference on Computer Vision (ECCV)},
  year={2020}
 }

@article{zhang2023vln,
  title={Vln-trans: Translator for the vision and language navigation agent},
  author={Zhang, Yue and Kordjamshidi, Parisa},
  journal={arXiv preprint arXiv:2302.09230},
  year={2023}
}

@inproceedings{wang2023lana,
  title={Lana: A language-capable navigator for instruction following and generation},
  author={Wang, Xiaohan and Wang, Wenguan and Shao, Jiayi and Yang, Yi},
  booktitle={Proceedings of the IEEE/CVF conference on computer vision and pattern recognition},
  pages={19048--19058},
  year={2023}
}

@inproceedings{kong2024controllable,
  title={Controllable navigation instruction generation with chain of thought prompting},
  author={Kong, Xianghao and Chen, Jinyu and Wang, Wenguan and Su, Hang and Hu, Xiaolin and Yang, Yi and Liu, Si},
  booktitle={European Conference on Computer Vision},
  pages={37--54},
  year={2024},
  organization={Springer}
}

\appendix

\section{Appendix}
\label{sec:appendix}

\subsection{Implementation Details}

%We include our implementation code in the supplementary materials of the submission and describe the relevant implementation details in this section. Upon acceptance, we will publish the codebase and datasets to support future research.

\textbf{Dataset Generation Pipeline}. We use \texttt{Gemini 1.5 Flash} \cite{georgiev2024gemini} as the landmark extractor to classify instructions into two categories: with landmark and without landmark. This classification allows us to apply different exploration strategies accordingly.  We use \texttt{gpt-3.5-turbo}\cite{openai2023gpt4} as the {Rewriter} to revise REVERIE instructions by replacing the original target object with a new object that does not exist in the corresponding room. To ensure that the selected object is truly absent, we employ \texttt{GLIP SWIN-Large}\cite{li2022grounded} as the {Verifier}, using a confidence threshold of 0.7. 

We provide below the prompt used for the rewriter(gpt-3.5-turbo):

\begin{lstlisting}[language=, caption={Prompt for the Rewriter}]
You should find a new target object to replace the old target_object and return me a new instruction.
Notice: the new target object must doesn't look like any objects(should be different type) in avoid_objects list.
Sometimes, you should review your answer and change the verb to which is suitable for the new target objects.
Important: you can only replace the target object and the verb about it.

Example:
inputs:
{
    'instruction': 'Go to bedroom at the back left side of the house and turn on the lamp nearest the bedroom door',
    'target_object': 'lamp',
    'avoid_objects': ['window', 'lamp', 'picture', 'bed'],
}
outputs:
{
    'new_target_object': 'the mirror',
    'new_instruction': 'Go to bedroom at the back left side of the house and take the mirror nearest the bedroom door',
}

explanation:
First, Choose 'the mirror' as the new target object because it isn't in the 'avoid_objects' list, and 'take' is a good verb for 'the mirror'.
So the new instructions is 'Go to bedroom at the back left side of the house and take the mirror nearest the bedroom door.'

Now it is your turn:
inputs:
    ___inputs___
outputs:

\end{lstlisting}

Below is an example from our dataset generated using our generation pipeline:

\begin{lstlisting}[language=, caption={Example for Our Dataset}]
{
    'original_instruction': 'Go to the laundryroom off of the garage and turn off the exhaust fan',
    'original_target_object': 'exhaust fan', 
    'avoid_objects': ['cabinet', 'fan', 'counter', 'trash can', 'floor', 'vent', 'washing machine', 'excercise equipment', 'faucet', 'dryer', 'rug', 'sink', 'roof', 'ceiling inset for fan', 'plant']

    'new_target_object': 'the stool',
    'new_instruction': 'Go to the laundryroom off of the garage and sit on the stool'
}
\end{lstlisting}

\textbf{\OurModel{} Framework}. In the Room-Object Extractor, we use \texttt{Gemini 1.5 Flash} \cite{georgiev2024gemini}. We first fine-tune DUET\cite{chen2022think} (the REVERIE version with object bounding box information) on \OurDataset{} using only the reach paths to obtain a Room Navigator. Notably, instead of using SPL to select the best model, we use Reach Success Rate (Reach SR) as the selection criterion. All other hyperparameters follow the default settings of DUET. 
At inference, we first run the Room Navigator from the start viewpoint. The Room Explorer is invoked only after the navigator enters the target room (i.e., $r(v_t)=r^\star$); otherwise, the episode terminates when the navigation step budget is exhausted. No ground-truth room entry or oracle hand-off is used at test time. 
%During inference, we assume that the Room Navigator has reached the target room. 
We then use \texttt{GPT-4o} or \texttt{GPT-3.5-turbo}\cite{openai2023gpt4} as the Room Explorer. This Room Explorer is adapted from NavGPT\cite{zhou2024navgpt}, with prompts modified to suit the \OurDataset{} task. Additionally, we incorporate distance information estimated by our FREE module. For the grounding component, we use the \texttt{grounding-dino-base}\cite{liu2024grounding} with a threshold of 0.75 to detect objects in each observation step taken by the Room Explorer.

Below is the prompt used for GPT-3.5-Turbo and GPT-4o in our experiments:

\begin{lstlisting}[language=, caption={Prompt for the In-Room Explorer}]
As an intelligent embodied agent, you will navigate in an indoor environment to reach a target viewpoint based on a given instruction, performing the Vision and Language Navigation (VLN) task.

The instruction will let you find all the target objects in a building.

But if you cannot find the target object, don't stop, keep exploring the whole room to find other objects, but you still should have a good strategy, don't waste time and energy to move.

You will move among static positions within a pre-defined graph, aiming for the nearest position to the object if the object is present.

You will receive a trajectory instruction at the start and will have access to step history (your Thought, Action, Action Input and Observation after the Begin! sign) and current viewpoint observation (including scene descriptions, objects, and navigable directions/distances within 3 meters) during navigation. Orientations range from -180 to 180 degrees, with 0 being forward, right 90 rightward, right/left 180 backward, and left 90 leftward.

And we will calculate how many meters extend in the direction of each viewpoint before hitting a wall. We hope this distance information can help you understand the spatial layout of the room. Please plan an effective exploration strategy based on this distance information.

For example, if I have 2 viewpoints to choose (A: 1m, B: 5m) but I cannot find the target object so I better choose viewpoint B because I may have more exploration space to find the target.

- Notice: You should have a good strategy to check whether the target object exists in the target room, and stop when you exploring all viewpoint in the target room.

Explore the environment while avoiding revisiting viewpoints by comparing current and previously visited IDs

If you think you are moving in circles, please stop and think whether any other objects may be hidden. If no, please output 'Final Answer: Not found'.

Continue by considering your location and the next viewpoint based on the instruction, using the action_maker tool.
And if you explored all the target room(no other viewpoint to move to), stop and output 'Final Answer: Not found!'.
Show your reasoning in the Thought section.

Follow the given format and use provided tools.
{tool_descriptions}
Do not fabricate nonexistent viewpoint IDs.

----
Starting below, you should follow this format, do not use other format:

Instruction: the instruction describing the whole trajectory
Initial Observation: the initial observation of the environment
Thought: you should always think about what to do next and why
Action: the action to take, must be one of the tools [{tool_names}]
Action Input: "Viewpoint ID", you should not choose object name or others, please only output "Viewpoint ID"
Observation: the result of the action
... (this Thought/Action/Action Input/Observation can repeat N times)
Thought: I found my target object, but I should check whether any other objects may be hidden.

or

Thought: I checked that no objects are hidden, I can stop.
Final Answer: Not found!

----

Begin!
Instruction: {action_plan}
Initial Observation: {init_observation}
Thought: I should start navigation according to the instruction, {agent_scratchpad}"""
\end{lstlisting}

\begin{figure}
    \centering
    \includegraphics[width=1\linewidth]{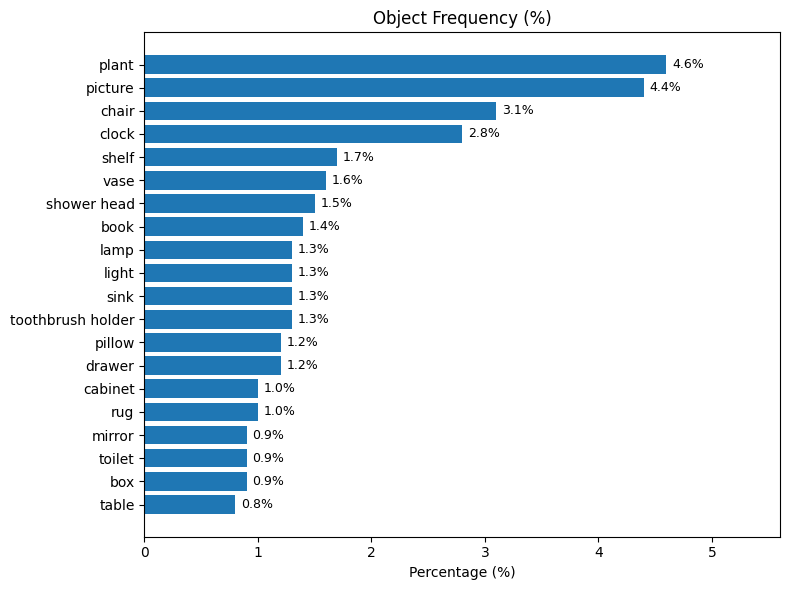}
\caption{Most occurred objects for \texttt{NOT-FOUND} instructions.}    \label{fig:figure4}
\end{figure}

\begin{figure}[t]
  \includegraphics[width=\columnwidth]{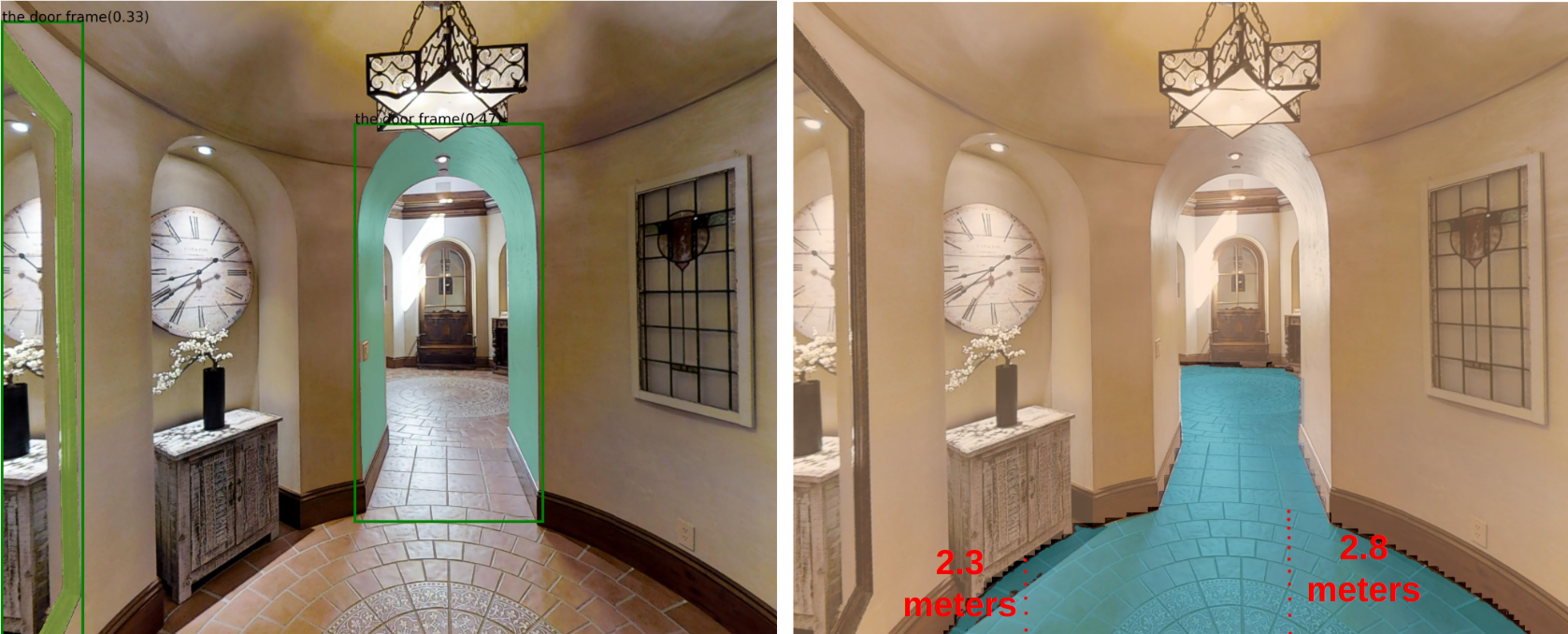}
  \caption{
  On the left, Grounding DINO detects the door frame. On the right, during raycasting, the process stops when encountering the door frame’s bounding box to avoid estimating incorrect distances (Note the red line indicating a distance of 2.8 meters).
 }
  \label{fig:figure5}
\end{figure}

\textbf{\OurModel{} Framework - Free-space Raycasting Estimation Engine
(FREE) }. We first utilize \texttt{Grounded SAM} (\texttt{GroundingDINO\_SwinT} + \texttt{sam\_vit\_h\_4b8939} version)\cite{ren2024grounded} with the prompt \textit{"the floor, the ground, the carpet"} to identify areas where the agent can move. We set the \texttt{box\_threshold} and \texttt{text\_threshold} parameters to 0.25 and 0.3, respectively. Additionally, our {FREE} module is designed to estimate the explorable distance within a single room. To avoid rays passing through doors and entering other rooms—potentially affecting decision-making—we further use the prompt \textit{"the door frame"} with \texttt{box\_threshold} set to 0.3. During raycasting, any intersection with a detected door frame is treated as a termination point for the ray. (Show in Figure \ref{fig:figure5})

After completing all segmentation, we perform Free-space Raycasting to estimate distances. First, we compute the vectors from the agent’s current viewpoint to each candidate viewpoint in the world coordinate system, then normalize these vectors to unit vectors with a length of 10 cm.

Next, we extend these unit vectors in the world coordinate system towards the candidate viewpoints. The points along these rays are projected into the camera coordinate system of the observation photos using the intrinsic and extrinsic camera parameters from Matterport3D. If the ray encounters an unwalkable area or a door frame bounding box, the extension stops.

Since each step corresponds to 10 cm, we estimate the maximum distance the agent can move in that direction by counting the number of valid steps. This process enriches the information available to the LLM for better environment understanding.

% \textbf{Baseline - DUET}. We adapt the DUET model to the \OurDataset{} task by extending its object prediction space to include an additional NOT-FOUND class. During inference, if the predicted probability for NOT-FOUND (i.e., the argmax) exceeds that of all other object or action candidates, the model outputs NOT-FOUND to indicate that the target is likely unreachable or absent.

% We follow the original DUET setup and retain all default hyperparameters. For model selection, we use Explore SPL as the evaluation metric to identify the best-performing checkpoint, as it accounts for both navigation success and the ability to avoid unnecessary exploration when the target cannot be found.

% \textbf{Baseline - NavGPT}. Only modify the original prompts

% \textbf{Baseline - Gemini}. Only modify the original prompts

\paragraph{Human Evaluation.}  We randomly sampled 5\% of the data. Annotators were shown: (1) the newly generated target object, and (2) all panoramas of the corresponding target room. They verified whether the new object truly did not exist in that room.

\subsection{Potential Risks}

We do not foresee direct risks related to malicious misuse, as our dataset \OurDataset{}{} does not include sensitive information, human identities, or unauthorized photos of private residences.

However, we acknowledge potential risks specific to our work. Since our dataset \OurDataset{} and framework \OurModel{} are designed to detect and respond to potentially incorrect instructions, improper handling of uncertainty may lead to premature task termination or refusal to act when the instruction is merely ambiguous rather than incorrect. Over-reliance on \texttt{NOT-FOUND} predictions could reduce system reliability in real-world deployment. To mitigate these risks, we recommend incorporating uncertainty estimation,  confidence calibration and fallback interaction (e.g., asking for clarification).

Fairness concerns such as overexposure to certain environments or instruction styles are also worth noting. As our dataset is built on REVERIE and Matterport3D Simulator, it primarily reflects a limited range of indoor environments. Additionally, all instructions are currently in English, which may disadvantage models trained for multilingual or non-English scenarios.

Overall, our work does not involve large-scale model training (e.g., LLM pretraining), does not pose privacy risks. We believe this benchmark promotes safer, more cautious navigation behavior by encouraging systems to recognize failure modes.

\subsection{License and Usage of REVERIE}

The REVERIE dataset used in this work is based on the Matterport3D environment and is distributed under a non-commercial research license. We follow all usage terms specified in the original REVERIE paper~\cite{qi2020reverie} and the Matterport3D dataset~\cite{chang2017matterport3d}. Our extensions (e.g., instruction augmentation and path generation) are built upon these publicly released annotations and simulation environments.

We confirm that our use of REVERIE complies with the dataset's license, and all modifications are intended for non-commercial academic research. The newly generated data will be released under similar terms to support future research in the community.

Furthermore, the datasets used do not contain any personally identifiable information or offensive content. We conducted manual sampling to verify this and ensure data quality. We have ensured that our use and extension of these artifacts respect the original creators' rights and adhere to ethical standards.

\subsection{Sensitivity to Minimum Reference-Coverage Threshold}
\label{app:refcov_sensitivity}

Our reference exploration protocol uses a minimum reference-coverage threshold
to filter episodes whose target-room visibility graph is too sparse for stable
evaluation.
In the main paper, we retain episodes with minimum reference coverage
$\geq 0.85$.
To verify that our conclusions are not sensitive to this choice, we evaluate
stricter retained subsets induced by thresholds $0.90$ and $0.95$, while
keeping agent trajectories unchanged.

\begin{table}[t]
\centering
\small
\setlength{\tabcolsep}{4pt}
\begin{tabular}{@{}lccc@{}}
\toprule
\textbf{th} & \textbf{ROAM} & \textbf{MapGPT} & \textbf{Retained} \\
 & \textbf{R\&D / REV} & \textbf{R\&D / REV} & \textbf{episodes} \\
\midrule
85\% & 37.6 / 6.1 & 14.0 / 3.2 & 100.0\% \\
90\% & 37.6 / 6.1 & 15.4 / 3.7 & 83.8\% \\
95\% & 37.8 / 6.5 & 14.2 / 3.3 & 76.1\% \\
\bottomrule
\end{tabular}
\caption{
Sensitivity to the minimum reference-coverage threshold.
Each cell reports Reach\&Decision SR / REV-SPL (both in \%).
This analysis varies only the retained evaluation subset
(i.e., stricter evaluability filtering), while keeping agent trajectories unchanged.
The retained-episodes column is reported relative to the 0.85 retained set.
}
\label{tab:refcov_sensitivity}
\end{table}

\paragraph{Interpretation.}
The results are stable across stricter retained subsets.
ROAM (GPT-3.5) maintains nearly identical Reach\&Decision SR
(37.6--37.8\%) and similar REV-SPL (6.1--6.5\%) across thresholds,
while MapGPT also shows only limited variation
(R\&D SR 14.0--15.4\%, REV-SPL 3.2--3.7\%).
Crucially, the method ranking is unchanged under all thresholds
(\textsc{ROAM} $>$ \textsc{MapGPT}),
indicating that our conclusions are not sensitive to the specific 0.85 choice
and that the threshold does not create method-dependent advantages.
Increasing the threshold simply filters out additional low-visibility cases,
reducing the number of retained episodes, while leaving the qualitative
conclusions unchanged.

\subsection{Use of AI Assistants}

We used AI assistants during the research and writing process to improve writing clarity and presentation.
Specifically, ChatGPT was used for language polishing (e.g., grammar, rephrasing, and clarity edits),
and for suggesting code improvements. It was also used to draft or edit \emph{non-empirical schematic
figures} (e.g., pipeline diagrams/icons) for illustration purposes only.
We also used ChatGPT and NotebookLLM to help organize and summarize literature during the early stages of the related work review. 
\textbf{No AI-generated images, captions, or other synthetic content were used as experimental inputs or as evidence
for the reported results. All AI-assisted outputs were reviewed and edited by the authors, and all critical decisions regarding modeling,
implementation, and analysis were made by the authors.}
%; all quantitative plots and qualitative examples are derived from real data/experiments and produced by the authors.

\subsection{Filtering, Distribution Shift, and Scalability}
\label{app:filtering_shift_scaling}

We further analyze whether the strict filtering used in \OurDataset{} substantially changes the underlying language/object distribution inherited from REVERIE.
To separate the effect of filtering from the effect of NF construction itself, we compare:
(i) \textbf{REVERIE vs.\ retained REVERIE} (filtering only), and
(ii) \textbf{REVERIE vs.\ final \OurDataset{}} (post rewrite/verify).
\paragraph{Interpretation.}
The filtering-only comparison shows extremely small distribution shift:
word/object rank agreement remains very high and normalized object proportions remain nearly unchanged.
This supports the view that strict filtering primarily enforces evaluability and label reliability, rather than introducing a substantial bias in language or object domain.
The additional shift from final \OurDataset{} is modest and expected, since NF construction intentionally replaces the original target with an absent but plausible alternative.
Importantly, this shift does not collapse the benchmark into a degenerate object distribution; rather, it reflects the intended transformation needed to create false-premise instructions.
\paragraph{Scalability.}
Although strict QC reduces the current benchmark size, the rewrite-and-verify pipeline is naturally scalable.
On the negative side, a single feasible episode can yield multiple NF variants by resampling alternative absent targets with blacklisting, and the difficulty of generated negatives can be controlled through prompting or sampling priors (e.g., same-category vs.\ distant-category substitutes, or landmark-cued vs.\ no-landmark settings).
On the positive side, feasible instances can be expanded through lightweight heuristics such as instruction paraphrases or start-node variations that preserve target-room reachability, and then paired with the same rewrite-and-verify pipeline to generate additional NF counterparts.
These properties make \OurDataset{} a high-confidence benchmark in its current form while leaving a clear path for future scaling.

\begin{table}[t]
\centering
\small
\setlength{\tabcolsep}{4pt}
\begin{tabular}{@{}lcc@{}}
\toprule
\textbf{Metric} & \textbf{Filtering only} & \textbf{Final \OurDataset{}} \\
\midrule
Words: Jaccard@50 $\uparrow$ & 0.887 & 0.852 \\
Words: Spearman $\rho$@50 $\uparrow$ & 0.942 & 0.863 \\
Words: JS $\downarrow$ & 0.0106 & 0.0148 \\
Objects: Spearman $\rho$ $\uparrow$ & 0.985 & 0.859 \\
Objects: Cosine sim.\ $\uparrow$ & 0.998 & 0.958 \\
Objects: JS $\downarrow$ & 0.0017 & 0.0414 \\
Objects: TV $\downarrow$ & 0.0330 & 0.1627 \\
\bottomrule
\end{tabular}
\caption{
Distribution-shift analysis for strict filtering and final NF construction.
``Filtering only'' compares full REVERIE against the retained REVERIE subset after QC filtering.
``Final \OurDataset{}'' compares full REVERIE against the final post rewrite/verify benchmark.
Higher is better for Jaccard, Spearman, and cosine similarity; lower is better for JS and TV.
}
\label{tab:filtering_shift}
\end{table}

\subsection{Model Size And Budget}

\textbf{Baseline - DUET} The number of parameters is nearly identical to that of DUET. Training on an RTX 4090 took approximately 48 hours

\textbf{Baseline - NavGPT} Using GPT-4 to run inference on the val\_unseen split costs approximately less than \$8 USD.

\textbf{Baseline - MapGPT} Using GPT-4 to run inference on the val\_unseen split costs approximately \$ 215 USD.

\textbf{Baseline - SoM} Using Gemini-2.0 Flash to run inference on the val\_unseen split costs approximately \$ 180 USD.

\textbf{Baseline - \OurModel{}} We trained DUET as a Room Navigator. Training on an RTX 4090 took approximately 48 hours. During the Fine-grained stage, we used GPT-3.5 for inference on the val\_unseen split, which incurred a cost of approximately less than \$3 USD.

% \section{Example Appendix}
% \label{sec:appendix}

% This is an appendix.

\end{document}